\documentclass[conference]{IEEEtran}
\usepackage{cite}
\usepackage{amsmath,amssymb,amsfonts}
\usepackage{algorithm}
\usepackage{algorithmic}
\usepackage{graphicx}
\usepackage{textcomp}
\usepackage{xcolor}
\usepackage{booktabs}
\usepackage{url}
\usepackage{pifont}
\usepackage{tikz}
\usepackage{hyperref}

\usepackage{caption}
\newcommand{\tabnote}[1]{\par\vspace{3pt}\parbox{\linewidth}{\footnotesize #1}}
\usetikzlibrary{arrows.meta}
\definecolor{hnavy}{HTML}{1E2D40}
\definecolor{hblue}{HTML}{1A6FA8}
\definecolor{horange}{HTML}{C85A15}
\definecolor{hgreen}{HTML}{177A5A}
\definecolor{hred}{HTML}{B2182B}
\newcommand{\best}[1]{\textcolor{hred}{#1}}
\definecolor{hpos}{HTML}{B2182B}
\definecolor{hneg}{HTML}{2166AC}
\definecolor{hseq}{HTML}{08519C}
\newtheorem{theorem}{Theorem}
\newtheorem{proposition}{Proposition}
\newtheorem{example}{Example}

\renewcommand{\IEEEproofindentspace}{0pt}

\begin{document}
\bstctlcite{IEEEexample:BSTcontrol}

\title{Beyond Interaction Scores: Selecting Feature Interactions via Independent Replication}

\author{\IEEEauthorblockN{Jincheng Gao}
\IEEEauthorblockA{
HKUST(GZ)\\
jgao627@connect.hkust-gz.edu.cn}
\and
\IEEEauthorblockN{Shuo Sun}
\IEEEauthorblockA{
HKUST(GZ)\\
shuosun@hkust-gz.edu.cn}}

\maketitle

\begin{abstract}
Learning feature interactions is essential for accurate prediction and interpretable modeling, yet a high interaction score may reflect fitting noise rather than a reproducible pattern. Ranking methods score feature pairs but leave the cutoff to the user; testing methods set the cutoff but rely on repeated model refits or a model of the joint feature distribution. We present LENS, a model-agnostic framework for selecting feature interactions through independent replication. LENS trains two models with the same learning algorithm on disjoint data and accesses them only through predictions. For each feature pair, it removes main effects from the joint-effect surface to isolate an interaction pattern, ranks pairs by its strength, and tests whether the second model reproduces it at a meaningful strength. A single multiple-testing correction determines which pairs to retain, using only two model fits regardless of the number of features. We prove that a replication ratio alone can retain arbitrarily weak patterns shared by both fits, whereas a strength floor restores selection consistency, and we characterize when normalization in Friedman's $H^2$ reverses interaction rankings. On 16 synthetic designs, LENS matches leading model-specific ranking methods (e.g., TreeSHAP) and recovers 91\% of material interactions, against 45\% for the knockoff-based DIAMOND, without selecting false pairs; with 1,225 candidate pairs and no interactions, it makes substantially fewer false discoveries. On large-scale French motor insurance data, it identifies interactions documented in the actuarial literature and recovers 90\% of the deviance gain obtained using all pairs. On two credit datasets, LENS selects few pairs while maintaining or slightly improving predictive performance over the additive baseline.
\end{abstract}


\section{Introduction}
\label{sec:intro}
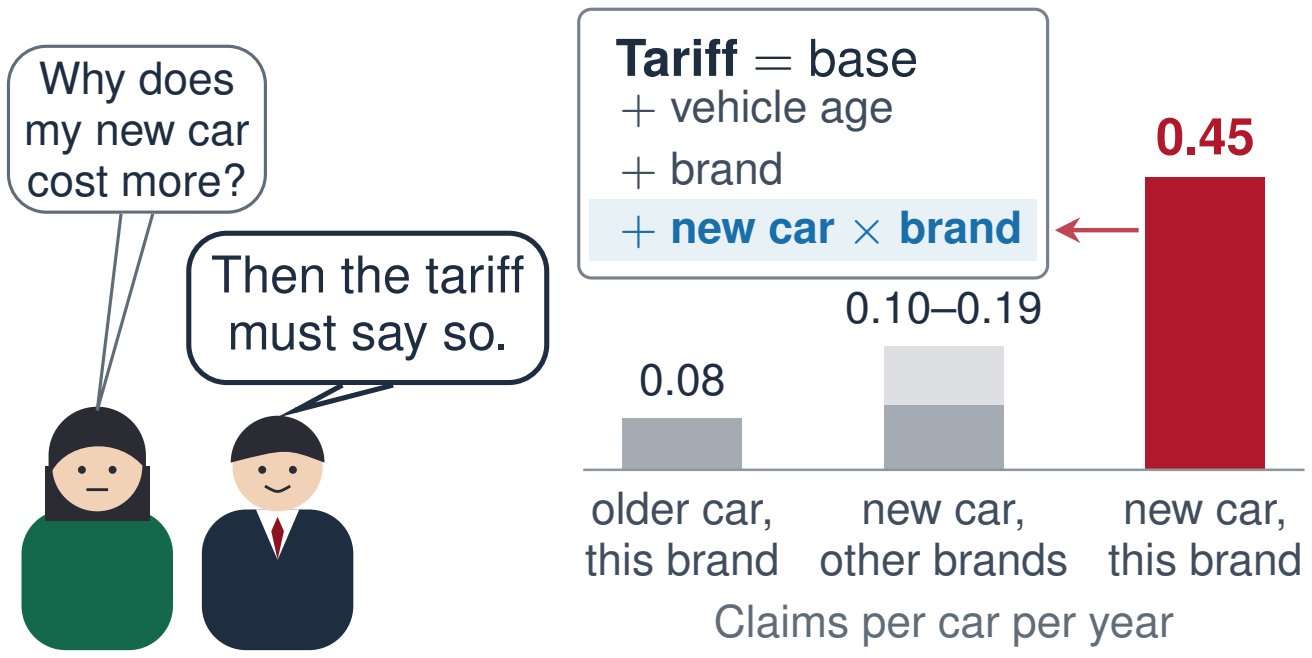
\begin{figure}[t]
  \centering
  \resizebox{\columnwidth}{!}{\begin{tikzpicture}[x=1cm, y=1cm, font=\sffamily\scriptsize,
  bubble/.style={draw=hnavy!70, fill=white, line width=0.5pt, rounded corners=6pt, align=center,
    font=\sffamily\scriptsize, text=hnavy, inner sep=3pt},
  thin/.style={-{Stealth[length=3pt, width=3pt]}, line width=0.6pt, hnavy},
  note/.style={font=\sffamily\scriptsize, text=hnavy!80, align=left}]
\definecolor{hskin}{HTML}{F1D2B6}
\definecolor{hhair}{HTML}{2B2B33}
\tikzset{
  actuary/.pic={
    \fill[hblue, rounded corners=7pt] (-0.42,0) rectangle (0.42,0.78);
    \fill[white] (-0.1,0.78) -- (0,0.62) -- (0.1,0.78) -- cycle;
    \fill[hskin] (0,1.02) circle (0.25);
    \fill[hhair] (-0.26,1.06) arc (180:0:0.26) -- (0.26,1.0) .. controls (0.1,1.14) and (-0.1,1.14) .. (-0.26,1.0) -- cycle;
    \fill[hhair] (-0.085,1.0) circle (0.022); \fill[hhair] (0.085,1.0) circle (0.022);
    \draw[hhair, line width=0.45pt] (-0.07,0.91) .. controls (0,0.95) .. (0.07,0.91);},
  regulator/.pic={
    \fill[hnavy, rounded corners=7pt] (-0.42,0) rectangle (0.42,0.78);
    \fill[white] (-0.12,0.78) -- (0,0.5) -- (0.12,0.78) -- cycle;
    \fill[hred!80!black] (0,0.74) -- (-0.035,0.66) -- (0,0.5) -- (0.035,0.66) -- cycle;
    \fill[hskin] (0,1.02) circle (0.25);
    \fill[hhair] (-0.26,1.04) arc (180:0:0.26) -- (0.26,1.04) .. controls (0.05,1.12) .. (-0.26,1.04) -- cycle;
    \fill[hhair] (-0.085,1.0) circle (0.022); \fill[hhair] (0.085,1.0) circle (0.022);
    \draw[hhair, line width=0.45pt] (-0.07,0.91) .. controls (0,0.87) .. (0.07,0.91);},
  customer/.pic={
    \fill[hgreen!85!black, rounded corners=7pt] (-0.42,0) rectangle (0.42,0.78);
    \fill[hhair] (-0.29,0.72) rectangle (0.29,1.02);
    \fill[hskin] (0,1.02) circle (0.25);
    \fill[hhair] (-0.27,1.08) arc (180:0:0.27) -- (0.27,1.0) .. controls (0.12,1.2) and (-0.2,1.16) .. (-0.27,0.95) -- cycle;
    \fill[hhair] (-0.085,1.0) circle (0.022); \fill[hhair] (0.085,1.0) circle (0.022);
    \draw[hhair, line width=0.45pt] (-0.06,0.89) -- (0.06,0.89);},
  car/.pic={
    \fill[pcol] (0,0.18) -- (0,0.45) -- (0.32,0.47) -- (0.5,0.72) -- (1.12,0.72) -- (1.34,0.47) -- (1.6,0.43) -- (1.6,0.18) -- cycle;
    \fill[white!85!pcol] (0.58,0.5) -- (0.66,0.66) -- (0.86,0.66) -- (0.86,0.5) -- cycle;
    \fill[white!85!pcol] (0.93,0.5) -- (0.93,0.66) -- (1.08,0.66) -- (1.22,0.5) -- cycle;
    \fill[hhair] (0.38,0.18) circle (0.15); \fill[white] (0.38,0.18) circle (0.06);
    \fill[hhair] (1.22,0.18) circle (0.15); \fill[white] (1.22,0.18) circle (0.06);}
}
\colorlet{pcol}{hblue}
\pic at (0.450,0) {customer};
\pic at (1.450,0) {regulator};
\node[bubble, anchor=south, font=\sffamily\scriptsize, inner sep=2.5pt] (bb) at (0.670,2.42) {Why does\\my new car\\cost more?};
\fill[white] ([xshift=-2.5pt]bb.south) -- (0.450,1.33) -- ([xshift=2.5pt]bb.south) -- cycle;
\draw[hnavy!70, line width=0.5pt] ([xshift=-2.5pt]bb.south) -- (0.450,1.33) -- ([xshift=2.5pt]bb.south);
\node[bubble, anchor=south, font=\sffamily\footnotesize, line width=0.8pt, draw=hnavy, inner sep=3.5pt] (rb) at (1.950,1.47) {Then the tariff\\must say so.};
\fill[white] ([xshift=-4pt]rb.south) -- (1.550,1.33) -- ([xshift=4pt]rb.south) -- cycle;
\draw[hnavy, line width=0.8pt] ([xshift=-4pt]rb.south) -- (1.550,1.33) -- ([xshift=4pt]rb.south);
\fill[hnavy!40] (3.365,1.000) rectangle +(0.66,0.288);
\node[font=\sffamily\scriptsize, anchor=south, text=hnavy] at (3.695,1.288) {0.08};
\node[font=\sffamily\scriptsize, text=hnavy!85, align=center, anchor=north] at (3.695,1.020) {\strut older car,\\\strut this brand};
\fill[hnavy!40] (4.815,1.000) rectangle +(0.66,0.360);
\fill[hnavy!15] (4.815,1.360) rectangle +(0.66,0.324);
\node[font=\sffamily\scriptsize, anchor=south, text=hnavy] at (5.145,1.684) {0.10--0.19};
\node[font=\sffamily\scriptsize, text=hnavy!85, align=center, anchor=north] at (5.145,1.020) {\strut new car,\\\strut other brands};
\fill[hred] (6.265,1.000) rectangle +(0.66,1.620);
\node[font=\sffamily\scriptsize, anchor=south, text=hred, font=\sffamily\footnotesize\bfseries] at (6.595,2.620) {0.45};
\node[font=\sffamily\scriptsize, text=hnavy!85, align=center, anchor=north] at (6.595,1.020) {\strut new car,\\\strut this brand};
\draw[hnavy!50, line width=0.4pt] (3.145,1.000) -- (7.145,1.000);
\node[font=\sffamily\scriptsize, text=hnavy!70] at (5.145,0.140) {Claims per car per year};
\draw[hnavy!60, fill=white, line width=0.5pt, rounded corners=2pt] (3.120,2.064) rectangle (5.720,3.554);
\node[font=\sffamily\footnotesize\bfseries, text=hnavy, anchor=north west] at (3.200,3.514) {Tariff \mdseries$=$ base};
\node[font=\sffamily\scriptsize, text=hnavy!85, anchor=west] at (3.220,2.989) {$+$ vehicle age};
\node[font=\sffamily\scriptsize, text=hnavy!85, anchor=west] at (3.220,2.659) {$+$ brand};
\fill[hblue!10] (3.180,2.159) rectangle (5.660,2.499);
\node[font=\sffamily\scriptsize, text=hblue, anchor=west] at (3.220,2.329) {$+$ \textbf{new car $\times$ brand}};
\draw[-{Stealth[length=4pt, width=4pt]}, hred!80, line width=0.7pt] (6.225,2.329) -- (5.760,2.329);
\end{tikzpicture}}
  \caption{Demonstration example on the importance of feature interaction in insurance pricing.}
  \label{fig:insurance}
\end{figure}

Two features interact when the effect of one depends on the value of the other. Learning insightful feature interactions is of vital importance for various forecasting and modeling tasks ~\cite{cheng2016wide,luo2019autocross,cordell2009detecting}. Fig.~\ref{fig:insurance} illustrates one example in insurance pricing, where the effect of vehicle age can vary across brands~\cite{schelldorfer2019nesting}. The elevated claim frequency associated with a particular combination of vehicle age and brand motivates an explicit interaction term in an otherwise additive tariff. Practitioners can use a flexible predictive model to identify candidate interactions, then add selected interactions as explicit terms in an interpretable model~\cite{lou2013accurate,schelldorfer2019nesting,havrylenko2024detection,lindholm2024blackbox}. The usefulness of this approach depends on choosing reliable interactions. Some patterns learned by the predictive model may arise from noise in its training sample and disappear when the model is trained on new data. This raises the question of which learned interactions are reproducible enough to retain.

Selecting these interactions is challenging because $p$ features give rise to $p(p-1)/2$ candidate pairs. Existing approaches fall into two broad groups, summarized in Fig.~\ref{fig:compare}. Ranking methods assign a score to each pair using model internals, such as tree structures or neural network weights~\cite{lundberg2020local,tsang2018detecting,janizek2021explaining}, or predictions alone~\cite{friedman2008predictive}. They prioritize candidates but leave users to choose a cutoff, often by retaining the top \(k\) pairs~\cite{tsang2018detecting,lindholm2024blackbox,henckaerts2022stakes}. A high score, however, can reflect fitting noise, so the ranking alone cannot determine which pairs have sufficient evidence for inclusion. Statistical testing methods address this selection problem, but typically require repeated model refits~\cite{friedman2008predictive,lampa2014identification,welchowski2022techniques,little2025iloco} or a model of the joint feature distribution~\cite{chen2025diamond}. These requirements incur substantial computational cost or introduce distributional modeling assumptions, especially when many candidate pairs must be considered.

\begin{figure*}[t]
  \centering
  \resizebox{\textwidth}{!}{\input{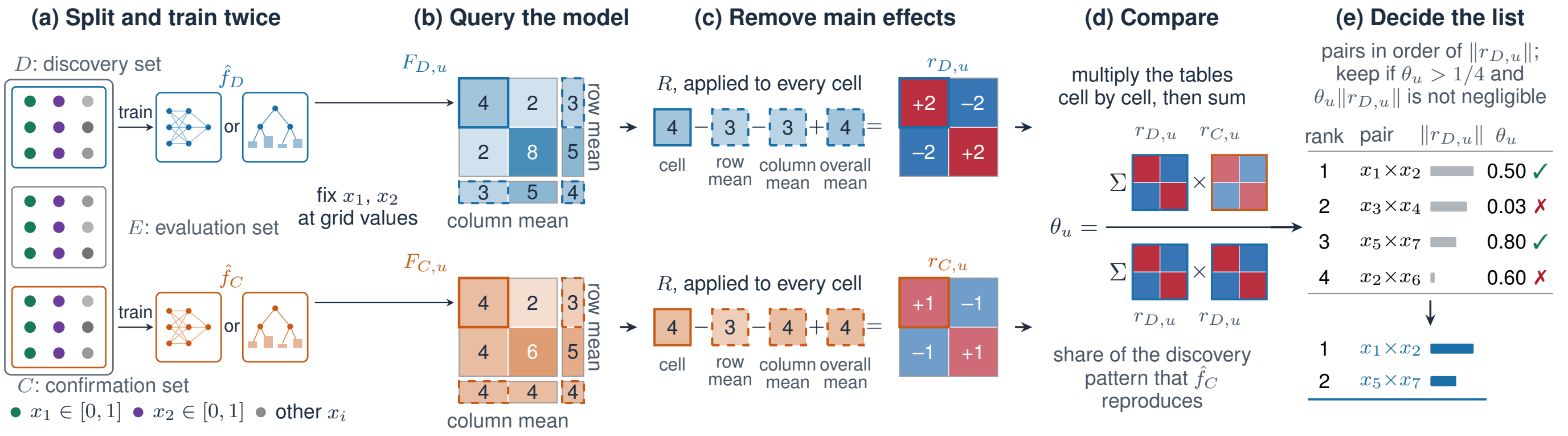}}
  \caption{Overview of LENS. (a) One data set is split into $D$, $C$, and
  $E$, and two models of any kind are trained on $D$ and $C$. (b) With $x_1$
  and $x_2$ fixed at grid values and the other features taken from real rows
  ($D$ for $\hat f_D$, $E$ for $\hat f_C$), the mean prediction fills the
  table $F_u$. (c) Removing the row and column means leaves the residual table
  $r_u$. (d) $\theta_u$ is the share of $r_{D,u}$ that $r_{C,u}$ reproduces.
  (e) Pairs ranked by $\|r_{D,u}\|$ are kept if the strength that $\hat f_C$
  reproduces, $\theta_u\|r_{D,u}\|$, is significantly above both
  $\|r_{D,u}\|/4$ and a floor set by the typical pair strength and the spread
  of the predictions, after one Benjamini--Yekutieli correction over all
  pairs; $x_3\times x_4$ is strong but not reproduced, and $x_2\times x_6$ is
  reproduced but negligible. Red
  positive, blue negative.}
  \label{fig:overview}
\end{figure*}

\begin{figure}[t]
  \centering
  \resizebox{\columnwidth}{!}{\begin{tikzpicture}[x=1cm, y=1cm, font=\sffamily\scriptsize,
  thin/.style={-{Stealth[length=3pt, width=3pt]}, line width=0.5pt, hnavy},
  note/.style={font=\sffamily\scriptsize, text=hnavy!80, align=left}]
\tikzset{
  minnet/.pic={
    \foreach \a in {0.2,0.5,0.8}{\foreach \b in {0.3,0.7}{\draw[pcol!60, line width=0.25pt] (0.22,\a) -- (0.55,\b);}}
    \foreach \b in {0.3,0.7}{\draw[pcol!60, line width=0.25pt] (0.55,\b) -- (0.82,0.5);}
    \foreach \a in {0.2,0.5,0.8}{\fill[pcol] (0.22,\a) circle (0.06);}
    \foreach \b in {0.3,0.7}{\fill[pcol] (0.55,\b) circle (0.06);}
    \fill[pcol] (0.82,0.5) circle (0.06);}
}
\fill[hnavy!4, rounded corners=4pt] (-0.05,3.07) rectangle (8.6,4.13);
\fill[hnavy!4, rounded corners=4pt] (-0.05,1.92) rectangle (8.6,2.99);
\fill[hnavy!4, rounded corners=4pt] (-0.05,0.77) rectangle (8.6,1.83);
\fill[hblue!7, rounded corners=4pt] (-0.05,-0.38) rectangle (8.6,0.68);
\node[font=\sffamily\scriptsize\bfseries, text=hnavy, anchor=south west] at (0.05,4.25) {Method};
\node[font=\sffamily\scriptsize\bfseries, text=hnavy, anchor=south] at (3.35,4.25) {Models trained};
\draw[hnavy!35, line width=0.5pt] (0,4.22) -- (8.6,4.22);
\node[font=\sffamily\footnotesize\bfseries, text=hnavy, anchor=west] at (0.05,3.80) {Ranking};
\node[text=hnavy!55, anchor=north west, align=left, inner sep=0pt] at (0.17,3.60) {$H^2$~\cite{friedman2008predictive}, TreeSHAP~\cite{lundberg2020local},\\NID~\cite{tsang2018detecting}};
\node[font=\sffamily\footnotesize\bfseries, text=hnavy, anchor=west] at (0.05,2.65) {Refit tests};
\node[text=hnavy!55, anchor=north west, align=left, inner sep=0pt] at (0.17,2.45) {permutation~\cite{lampa2014identification,welchowski2022techniques},\\iLOCO~\cite{little2025iloco}};
\node[font=\sffamily\footnotesize\bfseries, text=hnavy, anchor=west] at (0.05,1.43) {Knockoff tests};
\node[text=hnavy!55, anchor=west] at (0.05,1.13) {DIAMOND~\cite{chen2025diamond}};
\node[font=\sffamily\footnotesize\bfseries, text=hblue, anchor=west] at (0.05,0.15) {LENS};
\colorlet{pcol}{hnavy!55}
\draw[hnavy!55, fill=white, line width=0.5pt, rounded corners=1.5pt] (3.100,3.500) rectangle +(0.5,0.4);
\pic[xscale=0.5, yscale=0.4] at (3.100,3.500) {minnet};
\node[note, align=center] at (3.35,3.28) {one fit};
\colorlet{pcol}{hnavy!55}
\draw[hnavy!55, fill=white, line width=0.5pt, rounded corners=1.5pt] (3.110,2.510) rectangle +(0.5,0.4);
\pic[xscale=0.5, yscale=0.4] at (3.110,2.510) {minnet};
\colorlet{pcol}{hnavy!55}
\draw[hnavy!55, fill=white, line width=0.5pt, rounded corners=1.5pt] (3.030,2.430) rectangle +(0.5,0.4);
\pic[xscale=0.5, yscale=0.4] at (3.030,2.430) {minnet};
\colorlet{pcol}{hnavy!55}
\draw[hnavy!55, fill=white, line width=0.5pt, rounded corners=1.5pt] (2.950,2.350) rectangle +(0.5,0.4);
\pic[xscale=0.5, yscale=0.4] at (2.950,2.350) {minnet};
\node[note, align=center] at (3.35,2.13) {many refits};
\colorlet{pcol}{hnavy!55}
\draw[hnavy!55, fill=white, line width=0.5pt, rounded corners=1.5pt] (2.780,1.200) rectangle +(0.5,0.4);
\pic[xscale=0.5, yscale=0.4] at (2.780,1.200) {minnet};
\node[text=hnavy] at (3.45,1.40) {$+$};
\draw[hnavy!55, line width=0.5pt, rounded corners=1.5pt] (3.62,1.200) rectangle +(0.6,0.4);
\draw[hnavy!55, line width=0.6pt, smooth, domain=-1:1, samples=25, variable=\t] plot ({3.92+0.25*\t},{1.260+0.26*exp(-4*\t*\t)});
\node[note, align=center] at (3.35,0.98) {one fit + a model of $p(x)$};
\colorlet{pcol}{hblue}
\draw[hblue, fill=white, line width=0.5pt, rounded corners=1.5pt] (2.780,0.050) rectangle +(0.5,0.4);
\pic[xscale=0.5, yscale=0.4] at (2.780,0.050) {minnet};
\colorlet{pcol}{horange}
\draw[horange, fill=white, line width=0.5pt, rounded corners=1.5pt] (3.420,0.050) rectangle +(0.5,0.4);
\pic[xscale=0.5, yscale=0.4] at (3.420,0.050) {minnet};
\node[font=\sffamily\tiny, text=hblue] at (3.030,0.57) {$D$};
\node[font=\sffamily\tiny, text=horange] at (3.670,0.57) {$C$};
\node[note, align=center] at (3.35,-0.17) {two fits, any $p$};
\node[font=\sffamily\scriptsize\bfseries, text=hnavy, align=center, anchor=south] at (5.30,4.25) {Sets the\\list};
\node[font=\sffamily\scriptsize\bfseries, text=hnavy, align=center, anchor=south] at (6.60,4.25) {No model\\of $p(x)$};
\node[font=\sffamily\scriptsize\bfseries, text=hnavy, align=center, anchor=south] at (7.90,4.25) {Few\\fits};
\node at (5.30,3.60) {\footnotesize\color{hred}\ding{55}};
\node at (6.60,3.60) {\footnotesize\color{hgreen}\ding{51}};
\node at (7.90,3.60) {\footnotesize\color{hgreen}\ding{51}};
\node at (5.30,2.45) {\footnotesize\color{hgreen}\ding{51}};
\node at (6.60,2.45) {\footnotesize\color{hgreen}\ding{51}};
\node at (7.90,2.45) {\footnotesize\color{hred}\ding{55}};
\node at (5.30,1.30) {\footnotesize\color{hgreen}\ding{51}};
\node at (6.60,1.30) {\footnotesize\color{hred}\ding{55}};
\node at (7.90,1.30) {\footnotesize\color{hgreen}\ding{51}};
\node at (5.30,0.15) {\footnotesize\color{hgreen}\ding{51}};
\node at (6.60,0.15) {\footnotesize\color{hgreen}\ding{51}};
\node at (7.90,0.15) {\footnotesize\color{hgreen}\ding{51}};
\end{tikzpicture}}
  \caption{How LENS differs from existing methods. Sets the list: the method
  decides which pairs to keep. No model of $p(x)$: it needs no model of the
  joint law of the features. Few fits: the number of fits does not grow with
  the number of pairs.}
  \label{fig:compare}
\end{figure}

Independent replication offers a way to assess whether a learned interaction persists beyond a single training sample. If a pattern reflects a stable dependency captured by a learning algorithm, a model trained by the same algorithm on independent data should reproduce it. We build on this idea in LENS, a model-agnostic framework for selecting feature interactions through independent replication. LENS trains a discovery model and a confirmation model on disjoint data and accesses both only through predictions. It uses two model fits regardless of the number of features and performs the remaining computations through prediction queries.

Fig.~\ref{fig:overview} presents the procedure. For each feature pair, LENS constructs a joint-effect surface and removes the main effects of the two features to isolate their interaction pattern. The energy of the discovery pattern provides a score for ranking the pairs. Using a separate evaluation set, LENS then tests whether the confirmation model reproduces that pattern at a meaningful strength. A single multiple-testing correction determines which pairs to retain, potentially returning an empty set. The distinction between strength and replication is visible in the French motor insurance data. In the first split, VehAge$\times$VehBrand is the strongest pair in the discovery model, and the confirmation model reproduces 78\% of its pattern. VehAge$\times$Region ranks fifth, but only 3\% of its pattern is reproduced. LENS retains the former and rejects the latter.

Replication alone, however, is insufficient. Comparing independent fits removes the contribution of independent fitting variation in expectation, but both fits can share a systematic pattern induced by the learning algorithm. A replication ratio can consequently approach one even when the shared pattern is too weak to matter. We prove that a selection rule based on this ratio alone can retain arbitrarily weak shared patterns. LENS therefore also requires the reproduced interaction strength to exceed a floor. Under stated assumptions, this floor restores consistent selection of interactions whose limiting strength exceeds the threshold. Applying the Benjamini--Yekutieli correction~\cite{benjamini2001control} across all pairs further yields asymptotic false discovery rate control for the replication hypotheses under arbitrary dependence between pairs.

Our analysis also explains why the choice of ranking score matters. Friedman’s \(H^2\) divides interaction energy by the total variation of the joint-effect surface~\cite{friedman2008predictive}. This normalization makes a pair’s score depend on the magnitude of its main effects. A substantial interaction accompanied by strong main effects can receive a small score, while a weak interaction on a nearly constant surface can receive a large one. We characterize exactly when this normalization reverses the ranking of two pairs and establish consistency of the unnormalized interaction energy with a deterministic error bound. LENS uses this energy to rank candidates. On the same fitted XGBoost models, removing the normalization raises Recall@15 from 0.48 to 0.88, comparable to the tree-specific TreeSHAP score at 0.87. The ranking and replication test thus address complementary requirements of interaction discovery, quantifying the strength of a candidate and assessing whether an independent fit reproduces it sufficiently.

We evaluate LENS on 16 synthetic designs and three real-world datasets, covering diverse interaction structures, interaction-free settings, correlated features, and two model families. Its ranking accuracy matches leading model-specific methods. On the F1–F10 benchmarks, LENS with XGBoost recovers 91\% of material interactions, those that move predictions by at least 2\% of their spread, without selecting false pairs, compared with 45\% for DIAMOND with the same model family. In interaction-free settings with 1,225 candidate pairs, LENS makes substantially fewer false discoveries. On large-scale French motor insurance data, it identifies interactions documented in the actuarial literature and recovers 90\% of the deviance improvement obtained by adding all candidate pairs to the additive model. On two credit datasets, it selects few pairs and matches or slightly improves on the additive baseline. 
Our contributions are threefold:
\begin{itemize}
    \item \textbf{Interaction selection through independent replication.} We introduce LENS, a model-agnostic framework that ranks feature pairs and tests whether their interaction patterns replicate at a meaningful strength. It requires only two model fits and prediction access, with one multiple-testing correction determining the selected set.
    \item \textbf{Theory for replication and interaction ranking.} We prove that replication ratios alone can retain arbitrarily weak shared patterns and establish selection consistency with a strength floor under stated assumptions. We characterize ranking reversals caused by \(H^2\) normalization, establish consistency of the unnormalized score with a deterministic error bound, and prove asymptotic false discovery rate control for the replication hypotheses.
    \item \textbf{Evaluation of ranking, selection, and predictive utility.} Experiments on synthetic and real-world data demonstrate competitive ranking accuracy, recovery of material interactions with few false discoveries, and the predictive utility of selected interactions in additive models.
\end{itemize}
The code and every run behind the tables and figures are available at
\url{https://github.com/JinchengGao-Infty/LENS}.

\section{Problem Formulation}
\label{sec:formulation}

We are given $n$ rows with $p$ features and a learning algorithm, such as
gradient-boosted trees or a neural network, that fits a model
$\hat f:\mathbb R^p\to\mathbb R$. The output of $\hat f$ is on the link scale,
such as a log-rate or a logit. This section defines when a fitted model has a pair
interaction and states the problem.

\subsection{The interaction of a pair}
\label{sec:interaction}

A model takes all $p$ features at once, so the interaction of a pair
$u=(j,k)$ is defined on the model's joint effect of $x_j$ and $x_k$: its
partial-dependence surface~\cite{friedman2008predictive}
\begin{equation}
  F_u(a,b)=\mathbb E\bigl[\hat f(X\text{ with }X_j=a,\ X_k=b)\bigr],
  \label{eq:pd}
\end{equation}
the mean prediction when the two features are set to $a$ and $b$ and the other
features keep their distribution in the data. The pair has no interaction when
the surface is additive, $F_u(a,b)=c+\alpha(a)+\beta(b)$: the effect of one
feature does not depend on the other. Any departure from this
form is an interaction.

\noindent
\textbf{Example 1.} The left of Table~\ref{tab:example} is additive, with
$c=100$, $\alpha=(0,-10,-30)$, and $\beta=(0,20)$. Raising $x_2$ from 0 to 1 adds
20 in every row, so the effect of $x_2$ does not depend on $x_1$. The effects need
not be straight lines: $\alpha$ is not. The right of the table changes one cell.
Raising $x_2$ now adds 60 in the last row, so the effect of $x_2$ depends on $x_1$,
and no choice of $c$, $\alpha$, and $\beta$ can produce this table.

\begin{table}[t]
  \centering
  \caption{Example 1. Left: an additive table. Right: a table with an interaction. The columns $\delta$ give the effect of raising $x_2$ from 0 to 1.}
  \label{tab:example}
  \small
  \begin{tabular}{c ccc@{\hskip 14pt}ccc}
    \toprule
    & \multicolumn{3}{c}{Additive} & \multicolumn{3}{c}{Interaction} \\
    \cmidrule(lr){2-4}\cmidrule(lr){5-7}
    & $x_2=0$ & $x_2=1$ & $\delta$ & $x_2=0$ & $x_2=1$ & $\delta$ \\
    \midrule
    $x_1=1$ & 100 & 120 & 20 & 100 & 120 & 20 \\
    $x_1=2$ & 90 & 110 & 20 & 90 & 110 & 20 \\
    $x_1=3$ & 70 & 90 & 20 & 70 & \textbf{130} & \textbf{60} \\
    \bottomrule
  \end{tabular}
\end{table}

\subsection{The selection problem}
\label{sec:problem}

A fitted model has some interaction on nearly every pair, even when the data
have none, because it also fits noise. We call an interaction real when a model trained
by the same algorithm on independent rows learns it again, at a strength that
matters. A pair whose interaction is not real is a false discovery, and
$\mathcal H_0$ denotes the set of such pairs; Eq.~\eqref{eq:null} states this
hypothesis exactly, and Theorem~\ref{thm:fdr} controls errors on it. The target is thus defined through the learning algorithm:
an interaction learned on every sample is real for the model,
even when the generating function puts it on a correlated pair.

\noindent
\textbf{Problem.} Given the data and the learning algorithm, return a set
$\widehat G$ of the $P=p(p-1)/2$ pairs, ordered by interaction strength, such
that the false discovery rate
$\mathbb E\bigl[|\widehat G\cap\mathcal H_0|/\max(|\widehat G|,1)\bigr]$,
conditional on the fitted models, is asymptotically at most $\alpha$. The size
of $\widehat G$ is decided by the data, and $\widehat G$ may be empty. The
procedure should query the models only through their predictions and train a
number of models fixed in $p$.

\section{The LENS Method}
\label{sec:method}

LENS splits the data once into a discovery part $D$, a confirmation part
$C$, and an evaluation part $E$, and trains two models of the same kind (Fig.~\ref{fig:overview}(a)):
$\hat f_D$ on $D$ and $\hat f_C$ on $C$. The evaluation rows have no role in
training; they supply the contexts in Eq.~\eqref{eq:context-surface} for
the test. LENS then measures every pair's interaction in each model
(Section~\ref{sec:projection}), ranks the pairs by their strength in
$\hat f_D$, tests whether $\hat f_C$ reproduces them, and decides the list
with one correction over all pairs. Algorithm~\ref{alg:lens} summarizes the
procedure; Table~\ref{tab:notation} lists the notation.

\begin{table}[t]
  \centering
  \caption{Notation}
  \label{tab:notation}
  \footnotesize
  \setlength{\tabcolsep}{4pt}
  \begin{tabular}{@{}ll@{}}
    \toprule
    Symbol & Meaning \\
    \midrule
    $u=(j,k)$, $P$ & a pair of features; number of pairs, $p(p-1)/2$ \\
    $D$, $C$, $E$ & discovery, confirmation, and evaluation parts \\
    $F^{(i)}_u$, $F_u$ & context surface of row $i$; its average \\
    $R$ & projection that removes both main effects \\
    $r_{D,u}=RF_{D,u}$ & residual table of the discovery model \\
    $r_{C,u}=RF_{C,u}$ & residual table of the confirmation model \\
    $h_{C,u,i}=RF^{(i)}_{C,u}$ & residual table of the confirmation model in context $i$ \\
    $S_u$, $a_u$ & ranking score $\|r_{D,u}\|^2$; discovery strength $\|r_{D,u}\|$ \\
    $A_{u,i}$, $\theta_u$ & replication ratio in context $i$; its mean \\
    $\mu_u=\theta_ua_u$ & reproduced strength along the discovery pattern \\
    $s_D$, $\sigma_D$ & RMS strength over all pairs; spread of $\hat f_D$ \\
    $\eta$, $\kappa$, $\alpha$ & relative margin, scale floor, BY level \\
    $\Delta_u$, $M_u$ & interaction and main-effect energy of a surface \\
    \bottomrule
  \end{tabular}
\end{table}

\subsection{Measuring an interaction}
\label{sec:projection}

To compute Eq.~\eqref{eq:pd} for a pair $u=(j,k)$, LENS evaluates it on a grid of
values for each feature: quantiles with equal weights for a continuous feature,
levels weighted by their frequencies for a categorical one. Both are computed
from the features of $D$, before $\hat f_C$ is trained. Write $w_a$ and
$v_b$ for the weights of grid values $a$ and $b$. The context surface of row
$x_i$ is the table
\begin{equation}
  F^{(i)}_u(a,b)=\hat f\bigl(x_i\text{ with }x_j=a,\ x_k=b\bigr).
  \label{eq:context-surface}
\end{equation}
We set the two features to each pair of grid values and keep the rest
of the row. Averaging over rows gives the partial-dependence surface
$F_u=\frac1m\sum_{i=1}^{m}F^{(i)}_u$. Because the other features come from real
rows, their correlations are preserved.

This table still contains what each feature does alone. To keep only the
interaction, we remove these main effects: from every cell we subtract the mean
of its row and the mean of its column, and add back the overall mean,
\begin{equation}
  (RF)(a,b)=F(a,b)-\bar F(a,\cdot)-\bar F(\cdot,b)+\bar F(\cdot,\cdot),
  \label{eq:projection}
\end{equation}
where the means use the weights $w_a$ and $v_b$. The residual table $r_u=RF_u$
(Fig.~\ref{fig:overview}(c)) is the classical two-way interaction of the
surface. To measure it, write $\langle F,F'\rangle=\sum_{a,b}w_av_bF(a,b)F'(a,b)$
for the weighted average of the cell-by-cell products. The energy of the residual
table is $\Delta_u=\|r_u\|^2=\langle r_u,r_u\rangle$. An additive table $c+\alpha(a)+\beta(b)$ has
$R F=0$, so $\Delta_u=0$ exactly when the surface is additive on the grid, the
definition of Section~\ref{sec:interaction}.

\subsection{Ranking}

The larger the residual table, the stronger the interaction. For every pair
$u$, LENS computes the residual table of the discovery model,
$r_{D,u}=R F_{D,u}$, with contexts drawn from $D$, and ranks pairs by its energy,
\begin{equation}
  S_u=\|r_{D,u}\|^2 .
  \label{eq:score}
\end{equation}
It is the numerator of Friedman's $H^2$, computed on the grid rather than at
the data points. $H^2$ divides it by the variance of the whole surface;
Inglis et al.\ already drop this division~\cite{inglis2022visualizing}, and
Section~\ref{sec:theory} adds an error bound for the ranking and the exact
condition under which the division reverses it. The ranking orders the list; the test chooses
the pairs.

\subsection{The replication test}

A positive $S_u$ does not show that $\hat f_D$ learned something stable: a
model fitted to noise also has small interactions. LENS asks whether the pattern
$r_{D,u}$ reappears in the independently trained $\hat f_C$
(Fig.~\ref{fig:overview}(d)). For each evaluation row $i\in E$, let
$h_{C,u,i}=RF^{(i)}_{C,u}$ be the residual table of the confirmation model on
that row, and define
\begin{equation}
  A_{u,i}=\frac{\langle r_{D,u},h_{C,u,i}\rangle}{\|r_{D,u}\|^2},
  \qquad
  \theta_u=\mathbb E\bigl[A_{u,i}\bigr].
  \label{eq:ratio}
\end{equation}
$A_{u,i}$ compares the two tables cell by cell, and its mean $\theta_u$ measures
how much of the pattern of $\hat f_D$ the model $\hat f_C$ contains. It is 1 when
$\hat f_C$ reproduces the pattern exactly and 0 when $\hat f_C$ has nothing of it;
a different pattern in $\hat f_C$ does not count. It can also exceed 1 or turn
negative. Because $R$ and $\langle\cdot,\cdot\rangle$ are linear, the mean of
$A_{u,i}$ over the rows of $E$ equals $\langle r_{D,u},r_{C,u}\rangle/\|r_{D,u}\|^2$,
where $r_{C,u}=RF_{C,u}$ is the residual table of $\hat f_C$ with contexts from
$E$ (Fig.~\ref{fig:overview}(d)); the values per row are needed only for the
standard error of the test.

The two models are trained on disjoint data, so their fitting errors are
independent and cancel on average when the tables are compared. Only what both models
learned survives. A pattern that only one model learned gives $\theta_u$ near 0;
a pattern that both learned gives $\theta_u$ near the ratio of the two strengths.

A percentage, however, is blind to size. If both models carry the same faint
pattern, $\theta_u$ is near one however faint the pattern is, and
Section~\ref{sec:boundary} shows that a rule based on $\theta_u$ alone keeps such
pairs even as the pattern vanishes. LENS therefore also asks how large the
reproduced pattern is. Let $a_u=\|r_{D,u}\|$ be the strength of pair $u$ in
$\hat f_D$, and let $\mu_u=\theta_ua_u$ be the strength that $\hat f_C$
reproduces. LENS tests
\begin{equation}
  H_{0,u}:\ \mu_u\le\max(\eta a_u,\eta s_D,\kappa\sigma_D),
  \quad \eta=0.25,\ \kappa=0.02 ,
  \label{eq:null}
\end{equation}
where $s_D$ is the root-mean-square strength over all pairs and $\sigma_D$ is the
standard deviation of the predictions of $\hat f_D$. Table~\ref{tab:boundary}
gives the role of each term; all are fixed before $\hat f_C$ is trained.

\begin{table}[t]
  \centering
  \caption{Terms of the Boundary in Eq.~\eqref{eq:null}}
  \label{tab:boundary}
  \footnotesize
  \setlength{\tabcolsep}{4pt}
  \begin{tabular}{@{}lp{0.36\columnwidth}p{0.42\columnwidth}@{}}
    \toprule
    Term & A pair must reproduce & This rules out \\
    \midrule
    $\eta a_u$ & a quarter of its own strength & a strong pair that barely replicates \\
    $\eta s_D$ & a quarter of the typical strength & a weak pair that replicates a negligible pattern \\
    $\kappa\sigma_D$ & 2\% of the spread of the predictions & a model that learned no interaction, where every table shrinks \\
    \bottomrule
  \end{tabular}
\end{table}

To test this with the rows of $E$, divide by $a_u$: the hypothesis becomes
$\theta_u\le b_u$ with $b_u=\max(\eta a_u,\eta s_D,\kappa\sigma_D)/a_u$. Each row
gives one value $A_{u,i}$, so this is a one-sample test. With $\bar A_u$ and $s_u$
the mean and standard deviation of these values over the $m$ rows,
\begin{equation}
  Z_u=\frac{\sqrt m\,(\bar A_u-b_u)}{s_u},\qquad
  p_u=1-\Phi(Z_u).
  \label{eq:pvalue}
\end{equation}
Two cases need no division: a pair with $a_u=0$ has nothing to replicate and
gets $p_u=1$; if $s_u=0$, $p_u=\mathbf 1\{\bar A_u\le b_u\}$.

\noindent
\textbf{Example 2.} Take two pairs in the first split of the French motor
insurance data of Section~\ref{sec:exp-real}. On
the log-rate scale, the discovery model has $\sigma_D=0.591$ and
$s_D=0.040$, so $\eta s_D=0.010$ and $\kappa\sigma_D=0.012$, a 1.2\% change in
the claim rate.
VehAge$\times$VehBrand is the strongest pair, with $a_u=0.151$; its own term
dominates, and the boundary is $\eta a_u=0.038$. The confirmation model
reproduces $\mu_u=0.118$ along the discovery pattern ($\theta_u=0.78$), with
a standard error of $0.0002$ over 179,000 evaluation rows, so $Z_u\approx370$.
VehAge$\times$Region is the fifth strongest pair, with $a_u=0.060$ and
boundary $0.015$. The confirmation model reproduces $\mu_u=0.0019$
($\theta_u=0.03$), so $Z_u\approx-270$ and $p_u=1$. Both decisions lie far
from the boundary. With this many evaluation rows, the standard errors are
small, and the list is set by where the boundary lies rather than by the
level of the test (Section~\ref{sec:exp-sensitivity}).

\subsection{The decision: one correction over all pairs}

Testing many pairs at once lets some pass by chance, so LENS applies one
correction to the complete family. It computes $p_u$ for all $p(p-1)/2$ pairs
and applies the Benjamini--Yekutieli
procedure~\cite{benjamini2001control} at level $\alpha=0.05$. BY controls the false discovery rate
under arbitrary dependence between the p-values, which matters here because
all p-values share the same two models and the same evaluation rows. The
rejected pairs, ordered by $S_u$, form $\widehat G$. If no p-value passes,
$\widehat G=\varnothing$: LENS reports that the model has no interaction
that replicates.

\begin{algorithm}[t]
  \caption{LENS}
  \label{alg:lens}
  \begin{algorithmic}[1]
    \REQUIRE Data split into $D$, $C$, $E$; a learning algorithm; grids;
      margins $\eta$, $\kappa$; level $\alpha$.
    \STATE Train $\hat f_D$ on $D$.
    \FOR{every pair $u$}
      \STATE Compute $r_{D,u}=RF_{D,u}$ with contexts from $D$, and
        $S_u=\|r_{D,u}\|^2$.
    \ENDFOR
    \STATE Set $s_D=(P^{-1}\sum_uS_u)^{1/2}$, $\sigma_D$ the standard
      deviation of $\hat f_D$ over the contexts, and
      $b_u=\max(\eta a_u,\eta s_D,\kappa\sigma_D)/a_u$ with $a_u=\|r_{D,u}\|$.
    \STATE Freeze all $r_{D,u}$ and $b_u$. Train $\hat f_C$ on $C$.
    \FOR{every pair $u$}
      \STATE For each $i\in E$, compute $h_{C,u,i}=RF^{(i)}_{C,u}$ and
        $A_{u,i}$ (Eq.~\eqref{eq:ratio}).
      \STATE Compute $p_u$ (Eq.~\eqref{eq:pvalue}).
    \ENDFOR
    \STATE Apply BY at level $\alpha$ to $\{p_u\}$ over all pairs.
    \RETURN Rejected pairs, ordered by $S_u$; possibly none.
  \end{algorithmic}
\end{algorithm}

\subsection{Cost}
\label{sec:cost}

LENS trains two models, whatever the number of features. Everything else is
prediction. A pair with $G_j$ and $G_k$ grid values needs $G_jG_k$ predictions per context, so the ranking
costs $m_D\sum_uG_jG_k$ predictions and the test costs
$m_E\sum_uG_jG_k$, where $m_D$ and $m_E$ are the numbers of discovery and
evaluation contexts. With 12 grid points per feature, 50 features give 1,225
pairs and 144 grid cells per pair; the procedure takes about four
minutes on one GPU (Section~\ref{sec:exp-cost}). A refit test with empirical rank p-values
multiplies the training cost by the number of refits, which must
grow with $P$. After $B$ refits, a rank p-value cannot fall below $1/(B+1)$, while the
Benjamini--Yekutieli procedure at level 0.05 needs p-values below $0.05k/(mc_m)$
to keep $k$ of $m$ pairs, where $c_m=\sum_{i\le m}1/i$. So $B$ must exceed
$mc_m/(0.05k)-1$. Keeping three interactions among the 1,225 pairs of 50 features
takes about 63,000 refits: about five days on the same GPU, not four minutes.

\noindent
\textbf{Implementation.} LENS predicts all grid cells of a pair in one batched
call, and the pairs are independent, so they can be spread over devices. Each
context reduces to the scalar $A_{u,i}$, so the test stores only $m_E$ numbers per
pair. The procedure has no tuning loop: the grids, $\eta$, $\kappa$, and $\alpha$
are set once, and no step uses the outcome of the test to change an earlier step.

\section{Experiments on Synthetic Data}
\label{sec:experiments}

This section evaluates LENS on synthetic data with known interactions, and
Section~\ref{sec:real} applies it to real data. Section~\ref{sec:setup}
describes the models, designs, baselines, and metrics.
Section~\ref{sec:exp-ranking} first evaluates the residual energy as a ranking,
against scores built for one model family and against $H^2$.
Section~\ref{sec:exp-lists} then evaluates the lists that LENS returns,
including designs without interactions and with correlated features, as the
candidate pairs grow from 45 to 1,225. Section~\ref{sec:exp-sensitivity}
studies how the lists depend on the margins $\eta$ and $\kappa$ and on the level
$\alpha$, and Section~\ref{sec:exp-cost} reports the cost.

\subsection{Setup}
\label{sec:setup}

\noindent
\textbf{Models.} The main model is XGBoost with 1,000 trees, learning rate
0.05, and depth 6, with early stopping on 10\% of its training rows. To show
that LENS needs only predictions, we repeat the synthetic study with the
multilayer perceptron of DIAMOND~\cite{chen2025diamond}, which has layers of
140, 100, 60, and 20 ELU units. LENS splits the data into equal discovery,
confirmation, and evaluation parts, so each of its models sees a third of
the rows. It uses 12 grid points per feature, 3,000 discovery contexts, every
evaluation row as a test context, $\eta=0.25$, $\kappa=0.02$, and
$\alpha=0.05$. We fixed $\kappa$ on three pilot seeds that the results below
do not use.

\noindent
\textbf{Synthetic data.} Each design has $n=20{,}000$ rows of standard
normal features, Gaussian noise with standard deviation 0.25, and
$p\in\{10,30,50\}$ features, that is, 45, 435, or 1,225 candidate pairs. All
features beyond the active ones are irrelevant. \emph{Sparse} adds
$1.5x_0x_1+x_2x_3+0.75x_4x_5$ to the main effects
$\sin x_6+0.5x_7^2+0.5x_8$. \emph{Additive} keeps only these main effects.
\emph{Additive-corr} has main effects in $x_0$, $x_1$, and $x_4$, and makes
$x_2$ and $x_3$ noisy copies of $x_0$ and $x_1$ with correlation 0.9.
\emph{Pair-corr} has the interaction $x_0x_1$, small main effects, and the
same copies, so $(x_0,x_3)$, $(x_2,x_1)$, and $(x_2,x_3)$ are proxies of the
true pair and count as false. We also use F1--F10, the ten functions of the
NID and DIAMOND benchmarks~\cite{tsang2018detecting,chen2025diamond},
generated by DIAMOND's code with $p=30$, $n=20{,}000$, and no noise. Their true
pairs follow DIAMOND's convention, which splits higher-order terms into
pairs. All designs use 20 seeds.

\noindent
\textbf{Methods that decide.} We compare LENS with five methods that return a
list, each run with its published procedure.
DIAMOND~\cite{chen2025diamond} trains a model on the features and their
knockoffs and keeps the pairs that pass its knockoff filter at $q=0.2$. We
run its official code with its XGBoost model and with its MLP. Its lists barely
change at $q=0.05$ (FDP within 0.08), and on F1--F10, its own benchmark, its
FDP of 0.07--0.10 stays below $q$, as its paper reports. $H^2$ with a refit null follows Friedman and
Popescu~\cite{friedman2008predictive}: 100 refits on responses simulated from
an additive fit, then BY at 0.05 over all pairs~\cite{welchowski2022techniques}.
With rank p-values, BY kept no pair in any of the 440 runs, as the count in
Section~\ref{sec:cost} predicts, so we give $H^2$ a normal approximation to
its null distribution. iLOCO~\cite{little2025iloco} enters in both of its
versions. iLOCO-Split refits the model without $x_j$, without $x_k$, and
without both, for every pair; iLOCO-MP uses 10,000 minipatches of 20\% of the
rows and features. Both apply BY at 0.05 to normal p-values.
FAST~\cite{lou2013accurate} ranks the pairs, and the validation loss of the
resulting additive model with pair terms (GA$^2$M) chooses how many to keep from
$K\in\{0,1,2,3,5,8,12,20\}$.

\noindent
\textbf{Metrics.} A ranking is scored by Recall@15, the share of the true
pairs that rank in the top 15, and by average precision (AP), which
averages the precision at the rank of each true pair and so scores the whole
ranking. A list is scored against the generating function by its size, its
false discovery proportion (FDP), the share of its pairs that are false, and
its power, the share of the true pairs it contains. On the additive designs
every pair is false, so we report the share of runs that return any pair.
This FDP counts against the generating function, a different target from the
replication hypotheses of Theorem~\ref{thm:fdr}: a proxy that both fits learn
replicates but counts as false.
Each metric is averaged over seeds. On real data, where
the true pairs are unknown, a list is scored by the test loss of an additive
model using it (Section~\ref{sec:exp-real}).

\subsection{Ranking accuracy}
\label{sec:exp-ranking}

The residual energy ranks interactions as accurately as scores that read the
model's internals, and removing the denominator of $H^2$ is what makes the
difference. Table~\ref{tab:ranking} reports Recall@15 and AP over twelve functions with known interactions: Friedman's benchmark, a
smooth design, and F1--F10, each with $n=10{,}000$ and 32 seeds. The
model-specific scores run on the same fitted models as LENS. On XGBoost,
LENS reaches a Recall@15 of 0.88, against 0.87 for TreeSHAP. On a softplus
network it reaches 0.95, against 0.94 for Integrated Hessians and 0.67 for
NID. $H^2$, computed on the same XGBoost models and grids, reaches 0.48.

\begin{table}[t]
  \centering
  \caption{Ranking accuracy on twelve functions with known interactions. The last block uses reference implementations.}
  \label{tab:ranking}
  \small
  \begin{tabular}{llcc}
    \toprule
    Method & Model & Recall@15 & AP \\
    \midrule
    LENS & XGBoost & \textbf{0.88} & \textbf{0.90} \\
    TreeSHAP~\cite{lundberg2020local} & XGBoost & 0.87 & 0.88 \\
    $H^2$~\cite{friedman2008predictive} & XGBoost & 0.48 & 0.38 \\
    \midrule
    LENS & MLP & \textbf{0.95} & \textbf{0.96} \\
    Integrated Hessians~\cite{janizek2021explaining} & MLP & 0.94 & 0.95 \\
    NID~\cite{tsang2018detecting} & MLP & 0.67 & 0.62 \\
    \midrule
    TreeSHAP-IQ~\cite{muschalik2024beyond} & - & 0.87 & 0.87 \\
    Archipelago~\cite{tsang2020does} & - & 0.77 & 0.63 \\
    SHAP-IQ~\cite{fumagalli2023shapiq} & - & 0.72 & 0.71 \\
    Faith-Shap~\cite{tsai2023faith} & - & 0.56 & 0.49 \\
    hstats~\cite{mayer2023hstats} & - & 0.46 & 0.38 \\
    iRF~\cite{basu2018iterative} & - & 0.44 & 0.22 \\
    \bottomrule
  \end{tabular}
\end{table}

The failure of $H^2$ comes from its denominator: the stronger the main effects
of a pair, the lower $H^2$ ranks it (Proposition~\ref{prop:reversal}). We recorded both scores for every pair in 96
of these runs. Of the false pairs in the top 15 of $H^2$, 88\% join two
irrelevant features. Over 353,744 comparisons of a true pair with a false
pair from the same run, the energy orders 99.4\% correctly, and $H^2$
reverses 7.5\% of these. In the reversed comparisons, the median ratio
$M/\Delta$ is 736 for the true pair and 103 for the false one, which is
Eq.~\eqref{eq:reversal}. The true interactions sit on features with large
main effects, and the denominator dilutes them.

\begin{figure}[t]
  \centering
  \includegraphics[width=\columnwidth]{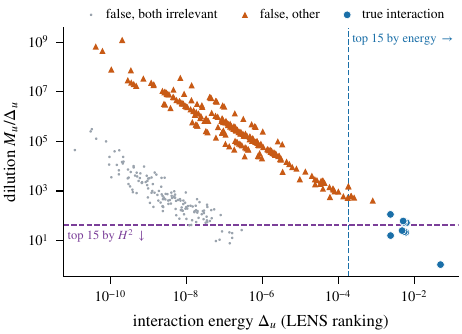}
  \caption{The two scores in one run of F4 ($p=30$, XGBoost). Each point is a
  pair; $H^2=1/(1+M_u/\Delta_u)$ ranks by height alone and the energy by
  horizontal position alone. Dashed lines mark the top 15 of each score. The
  71 pairs with $\Delta_u<10^{-11}$, which the model treats additively, are
  omitted; none is true.}
  \label{fig:h2}
\end{figure}

Fig.~\ref{fig:h2} shows the mechanism in one run of F4 with 435 pairs. Each
pair is placed by its energy $\Delta_u$, which orders the LENS ranking, and
by its dilution $M_u/\Delta_u$. Since $H^2=1/(1+M_u/\Delta_u)$, $H^2$ orders
the pairs by the vertical axis alone. The ten true pairs have the ten largest
energies, and the top 15 by energy contains all of them. They also carry
large main effects, with $M_u/\Delta_u$ up to 112, so $H^2$ admits only five
of them to its top 15. The other ten places go to pairs of two irrelevant
features. Their energies are $10^{-8}$ to $3\times10^{-7}$, four to five
orders of magnitude below the weakest true pair, but their faint surfaces
are as non-additive in shape as the true ones, with $M_u/\Delta_u$ between 8
and 42. The other false pairs contain at least one relevant feature. They
carry its main effect but almost no interaction, so they lie high on a line
of slope $-1$, below the true pairs in both scores.

\subsection{Deciding which pairs to keep}
\label{sec:exp-lists}

\begin{table*}[t]
  \centering
  \caption{Results on synthetic data, means over 20 seeds. The four designs use $p=50$ and F1--F10 use $p=30$. Red marks the best value in each column.}
  \label{tab:lists}
  \footnotesize
  \setlength{\tabcolsep}{4.2pt}
  \begin{tabular}{ll c c cc cc cc}
    \toprule
    & & Additive & Additive-corr & \multicolumn{2}{c}{Sparse}
    & \multicolumn{2}{c}{Pair-corr} & \multicolumn{2}{c}{F1--F10} \\
    \cmidrule(lr){3-3}\cmidrule(lr){4-4}\cmidrule(lr){5-6}\cmidrule(lr){7-8}\cmidrule(lr){9-10}
    Method & Model & FDP$\downarrow$ & FDP$\downarrow$ & FDP$\downarrow$ & power$\uparrow$
    & FDP$\downarrow$ & power$\uparrow$ & FDP$\downarrow$ & power$\uparrow$ \\
    \midrule
    LENS & XGBoost & \textbf{0.10} & \best{0.05} & \textbf{0.01} & \best{1.00} & \textbf{0.67} & \best{1.00} & \best{0.00} & \textbf{0.62} \\
    LENS & MLP & \best{0.00} & \textbf{0.15} & \best{0.00} & \best{1.00} & \best{0.03} & \best{1.00} & \best{0.00} & \textbf{0.66} \\
    DIAMOND & XGBoost & 0.90 & 0.90 & 0.18 & 0.93 & 0.35 & 0.55 & 0.07 & 0.31 \\
    DIAMOND & MLP & 0.65 & 0.90 & 0.14 & 0.93 & 0.12 & 0.20 & 0.10 & 0.29 \\
    $H^2$, normal refit null & XGBoost & 1.00 & 1.00 & 0.96 & \best{1.00} & 0.99 & \best{1.00} & 0.72 & \best{0.96} \\
    iLOCO-Split & XGBoost & 0.45 & 0.15 & 0.98 & \best{1.00} & 0.99 & 0.85 & 0.67 & 0.60 \\
    iLOCO-MP & trees & 1.00 & 1.00 & 0.99 & \best{1.00} & 1.00 & \best{1.00} & 0.99 & 0.30 \\
    FAST & EBM & 0.30 & 0.60 & 0.85 & \best{1.00} & 0.48 & \best{1.00} & 0.51 & 0.90 \\
    \bottomrule
  \end{tabular}
\end{table*}

\begin{figure}[t]
  \centering
  \includegraphics[width=\columnwidth]{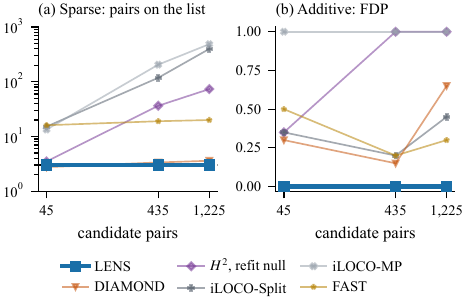}
  \caption{The list of LENS keeps its length as the candidate pairs grow from
  45 to 1,225; 20 seeds per point, with the same neural network for LENS and
  DIAMOND. (a) Mean list size on Sparse, which has three true pairs (dashed
  line). (b) FDP on Additive.}
  \label{fig:scaling}
\end{figure}

LENS is the only method whose lists are both complete and clean on the sparse
design, and it returns no false pair on the benchmark functions
(Table~\ref{tab:lists}). On the sparse
design it returns exactly the three true interactions in 19 of 20 runs, and
one extra pair in the twentieth; the same holds with 45 and 435 candidate
pairs. With the neural network, it returns exactly the three in all 60 runs.
DIAMOND finds 93\% of the true pairs, and about one pair in six on its lists
is false. The refit and leave-out tests find all three pairs but bury them:
the list of $H^2$ has 74 pairs and that of iLOCO-Split 399, and more than
95\% of each is false. FAST chooses its largest $K$ in every run.
These false pairs are not interactions of the model. Of those of $H^2$, 83\%
join two irrelevant features: their $H^2$ is a ratio of two near-zero
energies, high in some fits and low in others, so the normal approximation to
its null passes many of them (Section~\ref{sec:exp-ranking}). So do 67\% of
those of iLOCO-Split. It tests, as designed, the loss contrast of the refits
it has fitted, and for such a pair this contrast is a difference between four
refits, which is not zero; the differences between fits thus count as
interaction. FAST chooses $K$ by validation loss, which rewards any
pair term that lowers the loss. Of its false pairs, 83\% contain one feature
with a main effect, whose remaining curvature the pair term helps fit.
Fig.~\ref{fig:scaling} shows how the lists change as the candidate pairs grow
from 45 to 1,225. LENS keeps its list length; the refit
and leave-out tests do not. Appendix~\ref{app:synthetic} reports every design at every
number of features, for every method and decision rule.

Without interactions, LENS returns a pair in 7 of the 120 XGBoost runs on the
two additive designs, never more than one. With the neural network it
returns none on independent features, and one pair in 9 of 60 runs on
correlated features. DIAMOND with XGBoost returns pairs in 90\% of the runs at
$p=50$, and in 40\% already at $p=10$. In 96\% of its false pairs, both features have
main effects: the fitted model spreads a little of the nonlinear main effects
$\sin x_6$ and $0.5x_7^2$ over their pair, and the knockoff copies, which
carry no signal, cannot match it. The few false pairs of LENS on Additive
are of the same kind. $H^2$ and iLOCO-MP return pairs in
every run. For iLOCO-MP the leave-out score is not zero even when the function
is additive. Removing a feature changes how often every other feature
appears in the remaining minipatches, and with it the shrinkage of the
ensemble; even 100,000 minipatches did not shorten its lists.

\begin{figure}[t]
  \centering
  \includegraphics[width=\columnwidth]{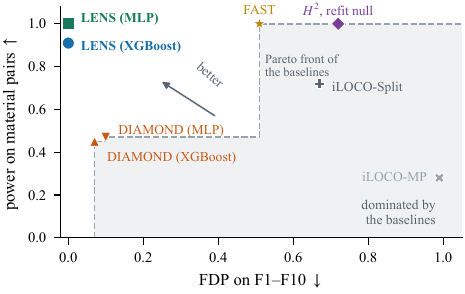}
  \caption{FDP and power on the material pairs of F1--F10, means over the ten
  functions and 20 seeds. The dashed line is the Pareto front of the
  baselines; LENS lies beyond it.}
  \label{fig:pareto}
\end{figure}

On F1--F10, LENS returns no false pair on any function with either model,
and its power is 0.62 with XGBoost and 0.66 with the network. The pairs it misses are weak. The benchmark labels a pair as true when its
features share a term of the formula, whatever the size of their interaction.
We computed the strength of every true pair from the generating function on the
same grids, in units of the standard deviation of the function. Of the 104 true
pairs, 40 fall below 2\%, and 38 of them come from splitting a higher-order term.
The other two have no interaction at all, because their terms are additive on
the unit cube. Against the
64 pairs above 2\%, LENS finds 91\% with XGBoost and all of them with the
network, while DIAMOND finds 45--47\% (Fig.~\ref{fig:pareto}). The 2\% cut
equals $\kappa$, but the gap does not depend on it: above 1\%, LENS finds
76--81\% and DIAMOND 38\%, and over all true pairs Table~\ref{tab:lists}
shows the same factor of two. Both LENS
models lie beyond the Pareto front of the baselines: every baseline has a
higher FDP, and none with an FDP below 0.5 finds more than half of the
material pairs. By strength, the cut is sharp. With the network, LENS detects every pair above
the floor and none below it; with XGBoost it misses some pairs just above the
floor. DIAMOND misses a third of the strongest pairs.

 FAST and $H^2$ find all material pairs, but
half or more of their lists are false. The scale floor certifies the
interactions that move predictions by at least 2\% of their spread, and the
weak pairs fall below it as intended.

The row $\kappa=0$ of Table~\ref{tab:sensitivity} shows why the scale floor is
needed. Without it, LENS on XGBoost finds every material pair of F1--F10, but
its FDP rises to 0.24, and on the additive designs it returns a list in every
run with 435 or more pairs. The neural network shows the failure of
Section~\ref{sec:boundary} in full. Its tables are smooth, so the two fits
reproduce each other's artifacts closely, and on the additive design the
relative terms keep 16, 146, and 403 pairs on average at $p=10$, 30, and 50. Yet on
F1--F10 the same rule is nearly exact, with an FDP of 0.02. The relative
terms judge each pair against the others, so they fail exactly when the
model has no interaction to compare with. The false pairs they keep on the
additive designs reproduce a median of 0.08\% (XGBoost) and 0.16\% (network)
of the prediction spread, and 99\% of them fall below the floor. The true
pairs of Sparse reproduce a median of 24\%. For the network, the scale floor
removes the failure and changes nothing on F1--F10 except the weak pairs.

The second model is needed as much as the floor. In the settings of
Table~\ref{tab:lists}, keeping the pairs whose
discovery strength $a_u$ clears the same floor, without $\hat f_C$, returns
about three false pairs in every additive XGBoost run, doubles the Sparse
list to 6.1 pairs (FDP 0.51), and raises the FDP on F1--F10 from 0 to 0.06,
at the same power. With the network it returns pairs in half of the
Additive-corr runs, against 15\% for LENS, and its Pair-corr FDP is 0.18,
against 0.03. Requiring both fits to clear the floor, without comparing their
tables, still returns pairs in every additive XGBoost run and gives a Sparse FDP
of 0.49: the second fit helps only through the comparison. The floor removes
what is too weak to matter, and replication what only one fit learned. Appendix~\ref{app:ablation} removes each ingredient
of LENS in turn, on synthetic and real data: without the comparison the false
pairs return on every design and the credit lists grow three- to fifteenfold at a
higher loss; the energy ranking cannot set the length of the list by itself; and
without the test the lists barely change at these sample sizes but lose the error
control of Theorem~\ref{thm:fdr}.

Correlated proxies remain the hard case for every method. On Pair-corr, LENS
with XGBoost returns the true pair with two proxies in 19 of 20 runs, an FDP
of 0.67. With the network it returns the true pair alone in 56 of 60 runs.
DIAMOND misses the true pair in 45--80\% of the runs, and 12--35\% of its
pairs are false. The lists of the refit and leave-out tests are almost
entirely false. As
Section~\ref{sec:boundary} explains, the XGBoost proxies are interactions of
the fitted model, reproduced at about a seventh of the true pair's strength,
so a test of the model's interactions correctly reports them.

The test itself is conservative. In a simulation that draws the ratios
$A_{u,i}$ directly, with every null pair at or below its boundary $b_u$, the
false discovery rate stays below 0.007 at a nominal level of 0.05 across 24
settings: 36 or 253 pairs, 128 to 2,048 contexts, Gaussian, $t_5$, and
log-normal noise, and independent and correlated pairs. Most of the slack comes
from BY, whose bound holds under any dependence.

\subsection{Sensitivity to the margins and the level}
\label{sec:exp-sensitivity}

The list depends on the scale floor $\kappa$, little on the relative margin
$\eta$, and hardly at all on the level $\alpha$. Table~\ref{tab:sensitivity}
re-decides every run under other settings. This needs no new fits, because
the reproduced strength $\hat\mu_u=a_u\bar A_u$ and its standard error do
not depend on the margins or the level; at the defaults the
recomputed lists match all 440 stored XGBoost lists.

\begin{table*}[t]
  \centering
  \caption{Sensitivity of LENS, one parameter at a time from the defaults ($\kappa=0.02$, $\eta=0.25$, $\alpha=0.05$). F1--F10 power counts the material pairs. Additive FDP pools $p=10,30,50$; Sparse and Pair-corr use $p=50$, with power 1.00 throughout.}
  \label{tab:sensitivity}
  \footnotesize
  \setlength{\tabcolsep}{4pt}
  \begin{tabular}{l cc cc cc cc cc rrr}
    \toprule
    & \multicolumn{2}{c}{F1--F10 FDP$\downarrow$} & \multicolumn{2}{c}{F1--F10 power$\uparrow$}
    & \multicolumn{2}{c}{Additive FDP$\downarrow$} & \multicolumn{2}{c}{Sparse FDP$\downarrow$}
    & \multicolumn{2}{c}{Pair-corr FDP$\downarrow$} & \multicolumn{3}{c}{List size} \\
    \cmidrule(lr){2-3}\cmidrule(lr){4-5}\cmidrule(lr){6-7}\cmidrule(lr){8-9}
    \cmidrule(lr){10-11}\cmidrule(lr){12-14}
    Setting & XGB & MLP & XGB & MLP & XGB & MLP & XGB & MLP & XGB & MLP
    & freMTPL2 & HELOC & UCI \\
    \midrule
    $\kappa=0$ & 0.24 & 0.02 & 1.00 & 1.00 & 0.93 & 1.00 & 0.54 & 0.20 & 0.78 & 0.90 & 13.0 & 23.2 & 35.2 \\
$\kappa=0.005$ & 0.06 & 0.00 & 1.00 & 1.00 & 0.49 & 0.97 & 0.28 & 0.20 & 0.71 & 0.90 & 13.0 & 19.0 & 33.0 \\
$\kappa=0.01$ & 0.00 & 0.00 & 1.00 & 1.00 & 0.33 & 0.45 & 0.09 & 0.00 & 0.70 & 0.36 & 13.0 & 10.4 & 17.0 \\
$\kappa=0.02$ (LENS) & 0.00 & 0.00 & 0.91 & 1.00 & 0.06 & 0.07 & 0.01 & 0.00 & 0.67 & 0.03 & 12.0 & 2.0 & 5.2 \\
$\kappa=0.03$ & 0.00 & 0.00 & 0.82 & 0.88 & 0.00 & 0.02 & 0.00 & 0.00 & 0.67 & 0.00 & 9.0 & 0.2 & 3.6 \\
$\kappa=0.05$ & 0.00 & 0.00 & 0.64 & 0.70 & 0.00 & 0.00 & 0.00 & 0.00 & 0.56 & 0.00 & 5.2 & 0.0 & 1.0 \\
$\kappa=0.1$ & 0.00 & 0.00 & 0.16 & 0.22 & 0.00 & 0.00 & 0.00 & 0.00 & 0.03 & 0.00 & 1.2 & 0.0 & 0.4 \\
\midrule
$\eta=0$ & 0.00 & 0.00 & 0.91 & 1.00 & 0.06 & 0.07 & 0.01 & 0.00 & 0.67 & 0.03 & 12.4 & 2.0 & 5.2 \\
$\eta=0.1$ & 0.00 & 0.00 & 0.91 & 1.00 & 0.06 & 0.07 & 0.01 & 0.00 & 0.67 & 0.03 & 12.4 & 2.0 & 5.2 \\
$\eta=0.5$ & 0.00 & 0.00 & 0.91 & 1.00 & 0.06 & 0.07 & 0.00 & 0.00 & 0.64 & 0.03 & 6.4 & 1.4 & 4.4 \\
$\alpha=0.01$ & 0.00 & 0.00 & 0.91 & 1.00 & 0.06 & 0.07 & 0.01 & 0.00 & 0.67 & 0.03 & 12.0 & 2.0 & 5.0 \\
$\alpha=0.2$ & 0.00 & 0.00 & 0.91 & 1.00 & 0.06 & 0.07 & 0.01 & 0.00 & 0.67 & 0.03 & 12.0 & 2.0 & 5.2 \\
\bottomrule

  \end{tabular}
\end{table*}

\begin{figure}[t]
  \centering
  \includegraphics[width=\columnwidth]{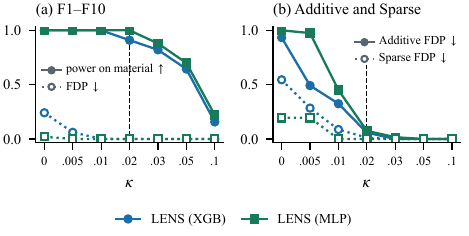}
  \caption{The scale floor trades power for clean null lists. Sparse uses
  $p=50$. The dashed line marks the default $\kappa=0.02$.}
  \label{fig:kappa}
\end{figure}

\noindent
\textbf{The scale floor.} The floor trades power for clean null lists, and the
trade is cheap because false and true pairs lie far apart: without the floor,
the false pairs reproduce about 0.1\% of the prediction spread and the true
pairs of Sparse 24\%. Below $\kappa=0.02$, the null designs leak. At
$\kappa=0.01$, XGBoost returns a pair in a third of the additive runs and the
network in 45\%, and the network keeps the proxies of Pair-corr. Above 0.02,
the lists stay clean but lose the weaker true pairs: power on the material
pairs of F1--F10 falls from 0.91 to 0.82 at $\kappa=0.03$ and to 0.64 at
0.05. The default is the smallest value in the grid at which both models
return a pair in fewer than one additive run in ten and recover the sparse
design exactly in at least 95\% of runs. We chose it on pilot seeds, and the
table shows that the choice carries over to the evaluation seeds. At $\kappa=0.1$, the XGBoost
proxies of Pair-corr fall below the floor as well, but so do 84\% of the
material pairs of F1--F10: a floor removes a proxy only by giving up the
true interactions that are as weak as the proxy. Fig.~\ref{fig:kappa} plots
the trade-off. On real data, the list shrinks steadily with $\kappa$, and on
the credit data, where the black box adds little, it is nearly empty
at 0.03.

\noindent
\textbf{The relative margin.} On the synthetic designs, $\eta$ between 0 and
0.5 changes almost nothing. Weak pairs face the scale floor, which is above
$\eta s_D$, and strong true pairs reproduce far more than half of their
strength. On freMTPL2, raising $\eta$ to 0.5 halves the list, from 12 pairs
to 6.4. There the real interactions reproduce between half and all of their
strength (Table~\ref{tab:fremtpl2}), so $\eta$ decides how complete a
replication must be. The default asks for a quarter, which lies on the flat part:
lowering $\eta$ to 0 changes the list by at most half a pair on average.

\noindent
\textbf{The level.} Changing $\alpha$ from 0.01 to 0.2 changes no synthetic
list, and on the real data it removes one pair from one split of UCI credit
at $\alpha=0.01$. Example~2 explains why: with thousands of evaluation rows,
the standard error of $\hat\mu_u$ is small compared with the distance of
most pairs from their boundary, so the decision is made by the hypothesis,
not by the level. With few evaluation rows the level matters, and the
simulation of Section~\ref{sec:exp-lists} shows that the test then stays
conservative, down to 128 rows. In practice, the choice that matters is where the boundary lies,
which is why LENS states it on the link scale, where an analyst can set it
from the size of an effect that matters to the application.

\subsection{Cost}
\label{sec:exp-cost}

Two fits are enough for a test over any number of pairs, and a refit test
with rank p-values cannot match this at any affordable number of fits. A refit test draws its
p-values from its refits, so they cannot fall below $1/(B+1)$. With 100
refits, BY kept no pair of $H^2$ in any of the 440 runs
(Section~\ref{sec:setup}). Each refit takes 7 seconds at $p=50$, so the
63,000 refits that Section~\ref{sec:cost} requires would take about five
days on one RTX 5090. LENS draws its p-values from the evaluation rows
instead, and two fits give p-values far below any refit floor: the strongest
pair on freMTPL2 has $Z_u\approx370$ (Example~2).

The time of LENS is set by prediction queries, which grow linearly with the
pairs and run in parallel. A run on Sparse takes 36, 63, and 252 seconds at
$p=10$, 30, and 50, so 1,225 pairs are tested in about four minutes. At
$p=50$ LENS makes about $1.7\times10^9$ predictions,
$1{,}225\times144\times9{,}600$, at about 7 million per second including both
fits. The refit methods are slower even with far too few refits for BY:
$H^2$ takes 35, 197, and 730 seconds, and iLOCO-Split, which refits for
every pair, 39, 667, and 1,838. Methods without a refit null finish in 92
(DIAMOND), 139 (iLOCO-MP), and 144 seconds (FAST) at $p=50$, and their lists
are those of Table~\ref{tab:lists}; DIAMOND also needs a generator for the
joint law of the features. The cost carries over to real data. On freMTPL2,
with 180,000 policies per part, a run takes 3.3--4.5 minutes, mostly for the
test of 179,000 evaluation rows; on credit data, under three.

\section{Experiments on Real Data}
\label{sec:real}
\label{sec:exp-real}

We apply LENS to one insurance dataset, where actuaries have reported
interactions, and to two credit datasets, where the black box adds little to an
additive model.

\subsection{Setup}
\label{sec:real-setup}

\noindent
\textbf{Data.} freMTPL2 has 678,013 French motor policies with nine rating
factors, so 36 pairs. We model claim counts with a Poisson XGBoost and an
offset of log exposure, and cap counts at 4 and exposure at 1, as is common
for this dataset; the four categorical factors use XGBoost's native splits.
LENS therefore tests interactions on the log claim rate, the scale on which
the tariff is additive. HELOC has 10,459 credit applications and UCI credit
has 30,000 card holders, each with 23 features and 253 pairs; both use a
logistic XGBoost. Each dataset is split five times. A fifth of the rows is
held out as a test set, and the rest is divided into the three parts of
LENS.

\noindent
\textbf{Measuring a list.} An additive model of boosted depth-one trees is
fitted on the discovery part. Each pair on a list enters as one term: its
discovery residual table, frozen as a lookup function, times a coefficient.
The coefficients are fitted jointly on the confirmation part, and we report
the loss on the test set. This is a generalized linear model (GLM) with fixed interaction shapes, a
deliberately simple use of the list; a GA$^2$M would refit the shapes.
We compare the LENS list with all pairs and with the top $K$ pairs by
energy for a fixed $K$, 7 on freMTPL2 and 10 on the credit data, which is
what a ranking without a test leaves to the user. With 20,000 policies the
LENS list has nearly this length (Section~\ref{sec:exp-insurance}).

\begin{table}[t]
  \centering
  \caption{Test loss on real data, mean over five splits: Poisson deviance on freMTPL2, log loss on credit data. Red: lowest loss among the lists. No floor: LENS with $\kappa=0$.}
  \label{tab:real}
  \footnotesize
  \setlength{\tabcolsep}{3.5pt}
  \begin{tabular}{l rc rc rc}
    \toprule
    & \multicolumn{2}{c}{freMTPL2} & \multicolumn{2}{c}{HELOC} & \multicolumn{2}{c}{UCI credit} \\
    \cmidrule(lr){2-3}\cmidrule(lr){4-5}\cmidrule(lr){6-7}
    List & pairs & loss$\downarrow$ & pairs & loss$\downarrow$ & pairs & loss$\downarrow$ \\
    \midrule
    None (additive) & 0 & 0.3128 & 0 & \best{0.5508} & 0 & 0.4312 \\
    LENS & 12.0 & \textbf{0.3095} & 2.0 & \best{0.5508} & 5.2 & \best{0.4303} \\
    No floor & 13.0 & 0.3095 & 23.2 & 0.5541 & 35.2 & 0.4318 \\
    Top $K$ by energy & 7 & 0.3108 & 10 & 0.5518 & 10 & \best{0.4303} \\
    All pairs & 36 & \best{0.3091} & 253 & 0.5915 & 253 & 0.4421 \\
    \midrule
    Black box & & 0.3055 & & 0.5633 & & 0.4325 \\
    \bottomrule
  \end{tabular}
\end{table}

\begin{figure}[t]
  \centering
  \includegraphics[width=\columnwidth]{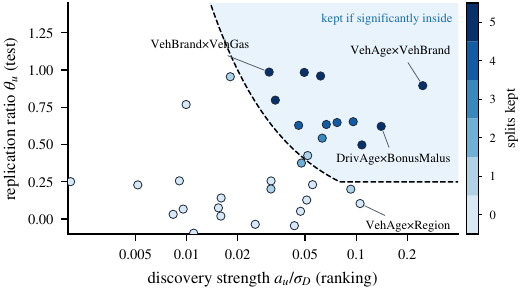}
  \caption{The 36 pairs of freMTPL2, averaged over five splits: discovery
  strength, which orders the ranking, against the replication ratio, which the
  test uses. The dashed line is the boundary of the null hypothesis,
  $\theta_u=\max(\eta,\kappa\sigma_D/a_u)$; here $\eta s_D$ lies below the
  scale floor. Color gives the number of splits in which LENS keeps the pair.}
  \label{fig:fremtpl2}
\end{figure}

\subsection{Insurance data}
\label{sec:exp-insurance}

\noindent
\textbf{Agreement with the actuarial literature.} The interactions LENS keeps
on freMTPL2 are those that actuaries have reported. Eleven pairs enter its
list in at least four of the five splits. They include all five interactions
that Schelldorfer and W\"uthrich identified~\cite{schelldorfer2019nesting}:
VehPower$\times$VehAge, VehPower$\times$VehBrand, VehAge$\times$VehBrand,
VehAge$\times$VehGas, and DrivAge$\times$BonusMalus. They also include all
five pairs to which the boosted model of Havrylenko and Heger gives a nonzero
$H$~\cite{havrylenko2024detection}, which adds DrivAge$\times$Region,
BonusMalus$\times$VehBrand, and VehAge$\times$BonusMalus. The two studies
share only two pairs, and the stable list contains the pairs of both. The remaining
three pairs, VehPower$\times$DrivAge, VehBrand$\times$VehGas, and
DrivAge$\times$VehBrand, appear in neither study. Table~\ref{tab:fremtpl2}
lists every pair that LENS keeps in some split. The stable pairs are not
simply the strongest: VehPower$\times$DrivAge and VehBrand$\times$VehGas rank
20th on average, but the confirmation model reproduces 80\% and 99\% of their
patterns, and both exceed the floor in every split.

\begin{table}[t]
  \centering
  \caption{Pairs kept on freMTPL2 in at least one of five splits, then the two strongest pairs never kept. Kept: splits that keep the pair; rank (by energy), $a_u/\sigma_D$, and $\theta_u$: means over splits; range: lowest and highest $\theta_u$. S\&W~\cite{schelldorfer2019nesting}, H\&H~\cite{havrylenko2024detection}: reported.}
  \label{tab:fremtpl2}
  \scriptsize
  \setlength{\tabcolsep}{2.5pt}
  \begin{tabular}{l c r c c c c c}
    \toprule
    Pair & kept & rank & $a_u/\sigma_D$ & $\theta_u$ & range & S\&W & H\&H \\
    \midrule
    VehAge$\times$VehBrand & 5 & 1.0 & 0.247 & 0.90 & 0.78 to 1.06 & \ding{51} &  \\
DrivAge$\times$BonusMalus & 5 & 2.2 & 0.141 & 0.62 & 0.43 to 0.89 & \ding{51} & \ding{51} \\
DrivAge$\times$Region & 5 & 4.4 & 0.108 & 0.50 & 0.31 to 0.67 &  & \ding{51} \\
VehAge$\times$VehGas & 5 & 11.4 & 0.062 & 0.96 & 0.47 to 1.45 & \ding{51} & \ding{51} \\
VehPower$\times$VehAge & 5 & 13.0 & 0.049 & 0.98 & 0.55 to 1.46 & \ding{51} &  \\
VehPower$\times$DrivAge & 5 & 19.6 & 0.033 & 0.80 & 0.54 to 1.35 &  &  \\
VehBrand$\times$VehGas & 5 & 20.0 & 0.031 & 0.99 & 0.59 to 1.59 &  &  \\
VehAge$\times$BonusMalus & 4 & 5.2 & 0.096 & 0.65 & 0.23 to 1.19 &  & \ding{51} \\
BonusMalus$\times$VehBrand & 4 & 8.6 & 0.077 & 0.65 & 0.08 to 1.00 &  & \ding{51} \\
DrivAge$\times$VehBrand & 4 & 9.2 & 0.067 & 0.63 & 0.22 to 1.20 &  &  \\
VehPower$\times$VehBrand & 4 & 14.6 & 0.046 & 0.63 & 0.41 to 0.96 & \ding{51} &  \\
VehAge$\times$DrivAge & 3 & 10.8 & 0.063 & 0.54 & 0.20 to 1.03 &  &  \\
DrivAge$\times$Density & 2 & 15.4 & 0.048 & 0.38 & $-$0.13 to 0.82 &  &  \\
BonusMalus$\times$Region & 1 & 5.4 & 0.093 & 0.20 & 0.00 to 0.44 &  &  \\
BonusMalus$\times$Density & 1 & 14.2 & 0.052 & 0.43 & $-$0.12 to 1.30 &  &  \\
VehAge$\times$Region & 0 & 4.6 & 0.105 & 0.10 & 0.03 to 0.21 &  &  \\
VehBrand$\times$Region & 0 & 11.8 & 0.055 & 0.23 & 0.12 to 0.33 &  &  \\
\bottomrule

  \end{tabular}
\end{table}

The test also explains the pairs that the literature disputes
(Fig.~\ref{fig:fremtpl2}). Havrylenko and
Heger's neural network ranks the pairs of BonusMalus and VehAge with the
regional factors high, and their AIC-based selection step recommends
BonusMalus$\times$Region~\cite{havrylenko2024detection}. Our discovery models
agree on the strength: VehAge$\times$Region is the fourth strongest pair on
average and is in the top five in four of five splits, so a top-5 cutoff,
as in the pipeline of Lindholm and Palmquist~\cite{lindholm2024blackbox},
would keep it.
The confirmation model, however, reproduces at most 21\% of its pattern, and
its table never significantly lowers the Poisson loss on held-out policies.
LENS never keeps it, and keeps BonusMalus$\times$Region in one split.

\noindent
\textbf{What the tables say.} The residual tables are themselves readable.
In the first split, the strongest interaction sits in one cell. New cars of brand B12, which make up
64\% of the new cars in the portfolio, have a residual of $+0.85$ on the log
rate, so their predicted claim rate is 2.3 times what the row and column
effects give. The raw data agree (Fig.~\ref{fig:insurance}): new B12 cars have 0.45 claims per
policy-year, against 0.08 for older B12 cars and 0.10 to 0.19 for new cars
of other brands. This one cell holds 64\% of the pair's energy, and the
confirmation model reproduces it at $+0.70$. DrivAge$\times$BonusMalus has the
shape that motivates this term in tariffs. In both tables, a bonus-malus
level of 95 raises the rate more for drivers under 36 than the additive
model allows, and less for drivers over 43, while the level 60 shows the
opposite trend. In VehAge$\times$BonusMalus, the bonus-malus scale is flatter
for new cars and steeper for cars older than 14 years, again in both tables.
The dropped pair looks different. VehAge$\times$Region puts 62\% of its
energy into the row of new cars, where the discovery model gives each region
its own correction, of either sign. The confirmation model has corrections of
the same size in that row, but they do not line up with the first: the two
tables agree in R21, a region with 0.4\% of the policies, and disagree in
most others, so the confirmation model reproduces 3\% of the pattern
($\theta_u=0.03$). Each model has
learned a region effect for new cars from the few hundred to few thousand
new cars per region in its part of the data, and the two effects differ. This is the pattern the test is designed to
reject, and an energy ranking alone cannot tell it apart from the kept pairs.

\begin{figure}[t]
  \centering
  \includegraphics[width=\columnwidth]{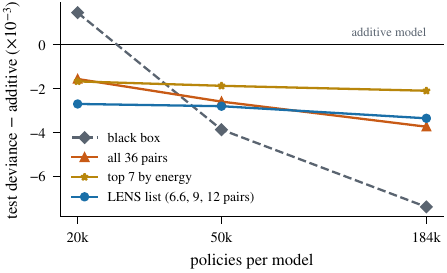}
  \caption{freMTPL2 test deviance relative to the additive model, by the
  number of policies used to fit each model (mean over five splits). Lower is
  better. The legend gives the mean length of the LENS list at each size.}
  \label{fig:sizes}
\end{figure}

\noindent
\textbf{Accuracy.} The list carries most of the gain of all pairs into the
additive model (Table~\ref{tab:real}). The 12 pairs that LENS keeps
on average close 45\% of the deviance gap between the additive model and the
black box, and 90\% of what all 36 pairs close (85--96\% across splits). The
7 strongest pairs close only 28\%, so the length that the test sets matters.
The value of
the list is clearest when data are scarce. With 50,000 policies per model,
the 9 pairs that LENS keeps match all 36 on average. With 20,000, the black box (0.3173) is
worse than the additive model (0.3159), yet the 6.6 pairs extracted from it
lower the deviance to 0.3132, below both and below all 36 pairs (0.3143), in
every split. Fig.~\ref{fig:sizes} shows the three sizes together. The black
box gains the most from more data, all 36 pairs gain steadily, and the LENS
list gains while staying short. With few policies the two fits can confirm
only the strongest interactions, so the list holds 6.6 pairs; with more, it
grows to 9 and 12. It grows by adding pairs: 73\% of the pairs kept with
20,000 policies are kept with all of them, and 82\% of those kept with
50,000. The test thus adapts the length of the list to the evidence, which a
fixed $K$ cannot do. The top 7 by energy does worse than the LENS list at
every size. At 20,000 policies the two lists have nearly the same length but
share only two pairs on average. The pairs that only the top 7 contains have
replication ratios between $-0.23$ and $0.23$; BonusMalus$\times$Region is
among the four strongest pairs in every split, yet its ratio never exceeds
0.23. With
little data, the strongest patterns of one fit are largely its own noise. At the same length, the
energy ranking alone is worse than LENS at every size and in every split
(Table~\ref{tab:app-real}).

\subsection{Credit data}
\label{sec:exp-credit}

When the black box has little to give, LENS returns a
short list, and the scale floor is what keeps it short. On both credit
datasets the black box is no better than the additive model. LENS keeps 2
pairs on HELOC and 5 on UCI credit. On HELOC the log loss is unchanged; on
UCI credit it drops by 0.0009, in four of five splits, and the additive model
with its pairs beats the black box. Without the floor, the test keeps 23 and
35 pairs, and these lists raise the log loss, on HELOC in every split. Here
the floor $\kappa\sigma_D$ is four to five times $\eta s_D$, whereas on
freMTPL2 it exceeds $\eta s_D$ by only 12--28\% and removes one pair on
average. The short lists vary across splits, with a Jaccard similarity of
0.05 and 0.21, as expected when no interaction is strong; only
PAY\_0$\times$BILL\_AMT1 on UCI credit enters in every split. The lists of
UCI credit concentrate on one feature: PAY\_0, the repayment status of the
latest month, appears in 8 of the 16
distinct pairs and in 16 of the 26 selections over five splits. On HELOC, the
two pairs kept twice both involve a summary risk score or the share of trades
never delinquent, ExternalRiskEstimate$\times$NumRevolvingTradesWBalance and
PercentTradesNeverDelq$\times$NetFractionRevolvingBurden. A short list that
changes from split to split is the honest answer when the interactions are
weak; the long lists without the floor change as much and add loss.

\section{Theory}
\label{sec:theory}

The experiments show what LENS does; this section explains why. All results
concern the interaction projection $R$ shared by the ranking and the test.
Section~\ref{sec:geometry} shows what $H^2$ measures and when it reverses
the order of two pairs, and that the energy ranks consistently while $H^2$
does not. Section~\ref{sec:cross} shows what the cross inner product of two
independent fits removes and what it keeps, and states the error control of
the replication test. Section~\ref{sec:boundary} shows why the test needs a
floor on the reproduced strength. All proofs are given in full.

\subsection{Why the energy ranks and $H^2$ does not}
\label{sec:geometry}
\label{sec:consistency}

Let $\mathcal A$ be the additive tables. Write a table as
$F=\mu+\alpha+\beta+r$, where
$\mu=\sum_{a,b}w_av_bF(a,b)$ is its weighted grand mean,
$\alpha(a)=\sum_bv_bF(a,b)-\mu$ and $\beta(b)=\sum_aw_aF(a,b)-\mu$ are its
centered weighted row and column means, and $r=RF$. Let
$M=\|\alpha\|^2+\|\beta\|^2$ be the main-effect energy and $\Delta=\|r\|^2$
the interaction energy.

\begin{proposition}[Decomposition]
\label{prop:decomposition}
The tables $\mu$, $\alpha$, $\beta$, and $r$ are pairwise orthogonal, $r$ is
orthogonal to every additive table, and $\|F-\mu\|^2=M+\Delta$. So $R$ is the
orthogonal projection onto the complement of $\mathcal A$, and
\begin{equation}
  \boxed{\;\Delta=\min_{A\in\mathcal A}\|F-A\|^2,\qquad
  H^2=\frac{\Delta}{M+\Delta}=\sin^2\vartheta,\;}
  \label{eq:energy}
\end{equation}
where $\vartheta$ is the angle between $F-\mu$ and the centered additive
tables.
\end{proposition}

\begin{IEEEproof}
Because the weights are a product, the inner product of a table that
depends only on $a$ with a table that depends only on $b$ factors into two
weighted means. Thus $\langle\alpha,\beta\rangle
=\bigl(\sum_aw_a\alpha(a)\bigr)\bigl(\sum_bv_b\beta(b)\bigr)=0$, and the
constant $\mu$ is orthogonal to $\alpha$, $\beta$, and $r$, each of which has
weighted mean zero. By Eq.~\eqref{eq:projection}, the weighted row means of
$r$ vanish:
$\sum_bv_br(a,b)=\sum_bv_bF(a,b)-\sum_{b'}v_{b'}F(a,b')-\mu+\mu=0$, and so do
its column means. Hence for any additive table $c+\alpha'(a)+\beta'(b)$,
$\langle r,\alpha'\rangle=\sum_aw_a\alpha'(a)\sum_bv_br(a,b)=0$,
and likewise $\langle r,\beta'\rangle=\langle r,c\rangle=0$. Since
$F-r=\mu+\alpha+\beta$ is additive and $r$ is orthogonal to all additive
tables, $r$ is the orthogonal projection of $F$ onto the complement of
$\mathcal A$, which gives the first identity in Eq.~\eqref{eq:energy}.
Pythagoras gives $\|F-\mu\|^2=\|\alpha\|^2+\|\beta\|^2+\|r\|^2$. Finally, the
projection of $F-\mu$ onto the centered additive tables is $\alpha+\beta$,
because the remainder $r$ is orthogonal to them, so
$\cos^2\vartheta=\|\alpha+\beta\|^2/\|F-\mu\|^2=M/(M+\Delta)$.
\end{IEEEproof}

\noindent\textbf{The energy is the distance of a surface to the nearest
additive table; $H^2$ is only the share of its variation that is not additive,
so it measures the direction of a surface and ignores its size.} A surface
with almost no variation can still point away from the additive tables and
score near one. The decomposition is exact because the grid carries product
weights; it does not require the features to be independent in the data,
since the background is averaged before the projection.

\begin{proposition}[Invariance to main effects]
\label{prop:invariance}
Let $\tilde f=\hat f+b_0+\sum_{\ell=1}^{p}b_\ell(x_\ell)$ for any functions
$b_\ell$. Replacing $\hat f_D$ or $\hat f_C$ by such an $\tilde f$ leaves
every ranking score $S_u$, the root-mean-square strength $s_D$, and every
ratio $A_{u,i}$ unchanged. In contrast, for a fixed residual $r\neq0$ and an
additive table $A_0$ with weighted mean zero,
$H^2(r+A_0)=\|r\|^2/(\|r\|^2+\|A_0\|^2)$, which tends to zero as $A_0$
grows.
\end{proposition}

\begin{IEEEproof}
Fix a pair $u=(j,k)$ and a context row $i$. The context surface of
$\tilde f$ is
\[
  \tilde F^{(i)}_u(a,b)=F^{(i)}_u(a,b)+b_j(a)+b_k(b)
  +b_0+\textstyle\sum_{\ell\ne j,k}b_\ell(x_{i\ell}).
\]
The added table is a function of $a$ plus a function of $b$ plus a
constant, so it lies in $\mathcal A$ and $R\tilde F^{(i)}_u=RF^{(i)}_u$. This
gives $\tilde h_{C,u,i}=h_{C,u,i}$ for every evaluation row. $R$ is linear,
so it commutes with the average over contexts, and
$\tilde r_{D,u}=r_{D,u}$. The scores $S_u$, the strengths $a_u$ and $s_D$, and
the ratios $A_{u,i}$ are functions of these tables alone. For the second
claim, $r+A_0$ has grand mean zero, its centered main effects sum to $A_0$,
and its residual is $r$; Proposition~\ref{prop:decomposition} gives
$M=\|A_0\|^2$ and $\Delta=\|r\|^2$.
\end{IEEEproof}

\noindent\textbf{Adding main effects changes no energy and no ratio
$A_{u,i}$, but can push any $H^2$ toward zero.} Adding an unrelated main
effect to the model therefore changes neither the order of the pairs nor any
relative threshold of the test. It moves only the floor $\kappa\sigma_D$,
which is measured in the spread of the predictions by design. The next result
gives the exact condition under which $H^2$ reverses two pairs.

\begin{proposition}[Reversal]
\label{prop:reversal}
If $\Delta_u>\Delta_v>0$, then $H^2$ ranks $v$ above $u$ if and only if
\begin{equation}
  \boxed{\;M_u/\Delta_u>M_v/\Delta_v .\;}
  \label{eq:reversal}
\end{equation}
\end{proposition}

\begin{IEEEproof}
By Proposition~\ref{prop:decomposition},
\[
  H^2_u-H^2_v
  =\frac{\Delta_uM_v-\Delta_vM_u}{(M_u+\Delta_u)(M_v+\Delta_v)} .
\]
The denominator is positive, so $H^2_v>H^2_u$ exactly when
$\Delta_vM_u>\Delta_uM_v$. Dividing by $\Delta_u\Delta_v>0$ gives
Eq.~\eqref{eq:reversal}.
\end{IEEEproof}

\noindent\textbf{$H^2$ ranks a weak interaction between features with small
main effects above a strong interaction between features with large main
effects.} If every pair had the same main-effect energy, $H^2$ and $\Delta$
would order the pairs identically; the reversal is caused by differences in
$M$ between pairs. The following example shows that this happens even for a
model that converges to the truth.

\begin{example}
\label{ex:four}
Let $x_1,\dots,x_4$ be independent and uniform on $\{-1,1\}$, let the true
function be $f^\star=10x_1+x_1x_2$, and let the fitted models be
$\hat f_n=f^\star+\varepsilon_nx_3x_4$ with $\varepsilon_n\downarrow0$.
Then $\hat f_n\to f^\star$ uniformly. With grids $\{-1,1\}$ and equal
weights, the surface of the true pair $(1,2)$ is $10a+ab$, because the
term $\varepsilon_nx_3x_4$ averages to zero, so $\Delta=1$, $M=100$, and
$H^2=1/101$. The surface of the spurious pair $(3,4)$ is $\varepsilon_ncd$,
because $10x_1+x_1x_2$ averages to zero, so $\Delta=\varepsilon_n^2$, $M=0$,
and $H^2=1$. For every $n$, $H^2$ ranks the spurious pair first, while the
energy ranks the true pair first once $\varepsilon_n<1$.
\end{example}

The root cause is that $H^2$ has no limit at the zero surface. For a unit
additive table $A$ and a unit interaction table $B$, $H^2(\epsilon A)=0$ and
$H^2(\epsilon B)=1$ for every $\epsilon>0$, although both tables tend to
zero. No value assigned to $H^2(0)$ makes it continuous there. A spurious
surface that vanishes as the fit improves therefore keeps the score its
shape dictates, whereas $\Delta(\epsilon B)=\epsilon^2$ vanishes.

Let $\mathcal S$ be the pairs whose target surface $F^\star_u$ is not additive
and $\mathcal N$ the others, and write $\hat F_u$ for the estimated surface;
the family is finite. A pair belongs to $\mathcal S$ only if its averaged
surface is non-additive: a term such as $x_jx_kx_\ell$ with $\mathbb Ex_\ell=0$
is non-additive in every context but averages out, which leaves $(j,k)$ in
$\mathcal N$.

\begin{theorem}[Energy ranks consistently]
\label{thm:energy}
Let $d=\min_{u\in\mathcal S}\|RF^\star_u\|$ and
$e=\max_u\|\hat F_u-F^\star_u\|$. Then
\begin{equation}
  \boxed{\;e<d/2\;\Longrightarrow\;
  \min_{u\in\mathcal S}\hat\Delta_u>\max_{v\in\mathcal N}\hat\Delta_v .\;}
  \label{eq:consistency}
\end{equation}
If $e\to0$ in probability, all interacting pairs rank above all
non-interacting pairs with probability tending to one, and Recall@$K\to1$
for every $K\ge|\mathcal S|$.
\end{theorem}

\begin{IEEEproof}
$R$ is an orthogonal projection, so $\|RG\|\le\|G\|$ for every table $G$.
By the reverse triangle inequality,
\[
  \bigl|\sqrt{\hat\Delta_u}-\|RF^\star_u\|\bigr|
  \le\|R(\hat F_u-F^\star_u)\|\le\|\hat F_u-F^\star_u\|\le e .
\]
For $v\in\mathcal N$, $RF^\star_v=0$, so $\hat\Delta_v\le e^2$. For
$u\in\mathcal S$, $\sqrt{\hat\Delta_u}\ge d-e>e$ when $e<d/2$, so
$\hat\Delta_u\ge(d-e)^2>e^2$. The family is finite and every pair in
$\mathcal S$ has a nonzero residual, so $d>0$, and $\Pr(e<d/2)\to1$ when
$e\to0$ in probability. On this event, the $|\mathcal S|$ highest energies
belong to $\mathcal S$, so every list of length $K\ge|\mathcal S|$ contains
all of $\mathcal S$.
\end{IEEEproof}

\noindent\textbf{Any estimate whose surfaces are within half of the weakest
true interaction ranks every interacting pair above every other pair.} The
bound is deterministic, so it holds in finite samples. Surface errors enter
the energy of a non-interacting pair only at second order. Nothing in the
bound requires the surfaces of non-interacting pairs to vanish: a pair with
large main effects and no interaction, such as $x_j+x_k$, has $RF^\star=0$ and
is covered. $H^2$ has no such guarantee. Let
$h^\star=\min_{u\in\mathcal S}\Delta^\star_u/(M^\star_u+\Delta^\star_u)$ be
the $H^2$ of the weakest interacting pair in the target.

\begin{proposition}[$H^2$ does not rank consistently]
\label{prop:h2}
Suppose the estimated surfaces converge on $\mathcal S$, and for some
$\delta,\pi>0$,
$\liminf\Pr\bigl(\max_{v\in\mathcal N}\hat H^2_v>h^\star+\delta\bigr)\ge\pi$.
Then the probability that $H^2$ ranks all interacting pairs above all
non-interacting pairs is at most $1-\pi$ in the limit.
\end{proposition}

\begin{IEEEproof}
For $u\in\mathcal S$, $\Delta^\star_u>0$, so the denominator
$M^\star_u+\Delta^\star_u$ is positive and $H^2$ is continuous at
$F^\star_u$. Hence $\hat H^2_u\to h^\star_u$ in probability, and over the
finite set $\mathcal S$, $\min_{u\in\mathcal S}\hat H^2_u\to h^\star$. Let
$C_m$ be the event that $H^2$ ranks correctly and $E_m$ the event in the
assumption. On $C_m\cap E_m$,
$\min_{\mathcal S}\hat H^2_u>\max_{\mathcal N}\hat H^2_v>h^\star+\delta$,
an event whose probability tends to zero. Therefore
$\Pr(C_m)\le\Pr(C_m\cap E_m)+1-\Pr(E_m)$, and
$\limsup\Pr(C_m)\le1-\pi$.
\end{IEEEproof}

By the scale argument above, the condition depends only on the shapes of the
spurious surfaces, not on their size. If a spurious surface is
$\varepsilon_nG$ for a fixed shape $G$ with $H^2(G)>h^\star+\delta$, the
condition holds however fast $\varepsilon_n$ vanishes, because
$H^2(\varepsilon_nG)=H^2(G)$; Example~\ref{ex:four} is this case with
$\pi=1$. Two forces make the condition easy to meet: main effects dilute the
true pairs and lower $h^\star$, and the number of pairs of irrelevant
features grows quadratically. Section~\ref{sec:exp-ranking} measured both on
fitted XGBoost models.

\subsection{What replication removes}
\label{sec:cross}
\label{sec:validity}

The ranking uses the self inner product of a table and the test uses the
cross inner product of two tables. The difference is what the second fit
adds. Let $c_u=\mathbb E[h_{C,u,i}]$ be the confirmation table averaged over
evaluation contexts, and let $\rho_D=\mathbb E[r_{D,u}\mid\mathcal G]$ and
$\rho_C=\mathbb E[c_u\mid\mathcal G]$ be the expected tables of the learner
given the grids and weights $\mathcal G$, over training samples and training
randomness. The grids are computed from $D$ before $\hat f_C$ is trained, so
given them the two fits remain independent.

\begin{proposition}[Cross products cancel variance, not bias]
\label{prop:cross}
If $\hat f_D$ and $\hat f_C$ are trained on independent samples with
independent randomness, then, with all expectations given $\mathcal G$,
\begin{equation}
  \boxed{\begin{aligned}
  \mathbb E\langle r_{D,u},c_u\rangle&=\langle\rho_D,\rho_C\rangle,\\
  \mathbb E\|r_{D,u}\|^2&=\|\rho_D\|^2+\mathbb E\|r_{D,u}-\rho_D\|^2 .
  \end{aligned}}
  \label{eq:cross}
\end{equation}
\end{proposition}

\begin{IEEEproof}
Given $\mathcal G$, the table $r_{D,u}$ is a function of the discovery part
and its contexts, and $c_u$ is a function of the confirmation fit and the
evaluation contexts. These are independent of $D$, and $\mathcal G$ is a
function of $D$, so given $\mathcal G$ the two tables are independent and the
expectation of each cell product factors:
$\mathbb E[r_{D,u}(a,b)c_u(a,b)\mid\mathcal G]=\rho_D(a,b)\rho_C(a,b)$.
The weights $w_av_b$ are fixed by $\mathcal G$, so summing gives the first
identity. The second is the bias--variance
decomposition of $\|r_{D,u}\|^2$, cell by cell.
\end{IEEEproof}

\noindent\textbf{Comparing two independent fits cancels their variance in
expectation but not the bias they share: independent fitting noise adds
nothing to the cross product on average, but a systematic pattern of the
learner does.} The self inner product carries the variance of the fit: a model
trained on data without any interaction still has
$\mathbb E\|r_{D,u}\|^2>0$, which is why every pair receives a positive score
from one model. If either learner has an expected table of zero, the cross
product has mean zero. What survives is the agreement of the two expected
tables. When the data contain an interaction $r^\star_u$ and the learner is
consistent, both expected tables approach $r^\star_u$, and the cross product
approaches $\|r^\star_u\|^2$. But a learner can have a nonzero expected table
where the data have no interaction, from the systematic part of its error,
and two fits of the same learner share it. Section~\ref{sec:boundary} shows
that this shared part is what a purely relative threshold cannot remove.

For the test, condition on everything fixed before the evaluation rows are
used: both fitted models, the grids, the tables $r_{D,u}$, and the thresholds
$b_u$. These are functions of the discovery and confirmation parts and of
constants fixed in advance. The ratios $A_{u,1},\dots,A_{u,m}$ of a pair are
then functions of independent evaluation rows.

\begin{theorem}[False discovery rate]
\label{thm:fdr}
If the $A_{u,i}$ are i.i.d.\ with finite positive variance, then at the
boundary $\theta_u=b_u$, $\sqrt m(\bar A_u-b_u)/s_u\to N(0,1)$ in
distribution, and for every $t\in(0,1)$,
$\limsup_{m\to\infty}\Pr(p_u\le t)\le t$ whenever $\theta_u\le b_u$.
Let the family contain $P$ pairs, $m_0$ of which satisfy $\theta_u\le b_u$,
and let $V$ and $R_{\mathrm{BY}}$ be the false and total rejections of BY at
level $\alpha$, with critical values $c_r=r\alpha/(PL_P)$ and
$L_P=\sum_{i=1}^P1/i$. For arbitrarily dependent p-values,
\begin{equation}
  \boxed{\;\mathbb E\!\left[\frac{V}{R_{\mathrm{BY}}\vee1}\right]
  \le\frac{\alpha m_0}{P}+m_0\varepsilon_m,\;}
  \label{eq:fdr}
\end{equation}
where $\varepsilon_m=\max_{u}\max_{r\le P}[\Pr(p_u\le c_r)-c_r]_+$ over the
null pairs, and $\varepsilon_m\to0$ as $m\to\infty$.
\end{theorem}

\begin{IEEEproof}
\emph{Pointwise validity.} Let $\sigma_u^2$ be the variance of $A_{u,i}$. The
central limit theorem gives $\sqrt m(\bar A_u-\theta_u)/\sigma_u\to N(0,1)$,
the law of large numbers gives $s_u\to\sigma_u$ in probability, and Slutsky's
theorem gives $\sqrt m(\bar A_u-\theta_u)/s_u\to N(0,1)$. At the boundary this
is $Z_u$, so $p_u=1-\Phi(Z_u)$ converges in distribution to a uniform variable
and $\Pr(p_u\le t)\to t$. Inside the null, $\theta_u<b_u$, write
\[
  Z_u=\frac{\sqrt m(\bar A_u-\theta_u)}{s_u}
  -\frac{\sqrt m(b_u-\theta_u)}{s_u}.
\]
The first term is bounded in probability and the second tends to $+\infty$,
so $Z_u\to-\infty$, $p_u\to1$ in probability, and $\Pr(p_u\le t)\to0$.

\emph{False discovery rate.} We follow the argument of Benjamini and
Yekutieli~\cite{benjamini2001control}, keeping the marginal excess. Set
$c_0=0$, $B_1=[0,c_1]$, and $B_r=(c_{r-1},c_r]$ for $r\ge2$. BY rejects the
$R_{\mathrm{BY}}$ smallest p-values, where $R_{\mathrm{BY}}$ is the largest
$r$ with $p_{(r)}\le c_r$. If a null pair $u$ is rejected and $p_u\in B_r$,
then $r\le R_{\mathrm{BY}}$. Hence, writing $F_u(t)=\Pr(p_u\le t)$,
\begin{align*}
  \mathbb E\!\left[\frac{\mathbf 1\{u\text{ rejected}\}}{R_{\mathrm{BY}}\vee1}\right]
  &\le\sum_{r=1}^{P}\frac{\Pr(p_u\in B_r)}{r}\\
  &=\sum_{r=1}^{P-1}\frac{F_u(c_r)}{r(r+1)}+\frac{F_u(c_P)}{P},
\end{align*}
where the equality is summation by parts. This step uses only the marginal
law of $p_u$, which is why any dependence between pairs is allowed. Bound
$F_u(c_r)\le c_r+\varepsilon_m$. The coefficients satisfy
$\sum_{r<P}c_r/(r(r+1))+c_P/P=\alpha(L_P-1+1)/(PL_P)=\alpha/P$ and
$\sum_{r<P}1/(r(r+1))+1/P=1$, so each null pair contributes at most
$\alpha/P+\varepsilon_m$. Summing over the $m_0$ null pairs gives
Eq.~\eqref{eq:fdr}. By the first part, $F_u(c_r)\to c_r$ for a null pair at
its boundary and $F_u(c_r)\to0$ for one inside its null; the family and its
critical values are finite, so $\varepsilon_m\to0$.
\end{IEEEproof}

\noindent\textbf{Conditional on the two fits, the replication test
asymptotically controls the false discovery rate under any dependence between
pairs.} Every p-value uses the same two models and evaluation rows, so this
dependence must be allowed. The guarantee is conditional on the two fits, so
it holds for the models that the user actually has, not on average over models
the user might have trained. The family is fixed while the number of
evaluation rows grows; a family that grows with the data would need uniform
bounds on the normal approximation, which we do not pursue.

\subsection{Where replication alone fails}
\label{sec:boundary}

Theorem~\ref{thm:fdr} holds for any threshold fixed before evaluation,
including the purely relative one, $b_u=\eta$. The threshold decides which
pairs the list contains as the fits improve.

\begin{proposition}[A floor gives consistent selection]
\label{prop:floor}
Suppose that for every pair $r_{D,u}$ and $c_u$ converge in probability to a
common table $r^\dagger_u$, possibly zero, and that $\bar A_u-\theta_u\to0$
and $s_u/\sqrt m\to0$ in probability.
(a) With the relative threshold $b_u=\eta<1$, LENS keeps every pair with
$r^\dagger_u\neq0$ with probability tending to one, however small
$\|r^\dagger_u\|$ is.
(b) With the threshold $\max(\eta a_u,\tau_u)$ on $\mu_u$, where $\tau_u$ is
fixed before evaluation and converges to $\tau^\dagger_u>0$, and no
$\|r^\dagger_u\|$ equals $\tau^\dagger_u$, LENS selects with probability
tending to one
\begin{equation}
  \boxed{\;\widehat G=\{u:\|r^\dagger_u\|>\tau^\dagger_u\}.\;}
  \label{eq:floor}
\end{equation}
The floor $\max(\eta s_D,\kappa\sigma_D)$ of Eq.~\eqref{eq:null} has a positive
limit whenever the limit model is not constant on the contexts, because
$\sigma_D$ converges to its standard deviation there.
\end{proposition}

\begin{IEEEproof}
(a) For a pair with $r^\dagger_u\ne0$, by the definition of $\theta_u$ and
the Cauchy--Schwarz inequality,
\[
  |\theta_u-1|=\frac{|\langle r_{D,u},c_u-r_{D,u}\rangle|}{\|r_{D,u}\|^2}
  \le\frac{\|c_u-r_{D,u}\|}{\|r_{D,u}\|}\to0,
\]
because the numerator tends to zero and the denominator to
$\|r^\dagger_u\|>0$. Then $\bar A_u-\eta\to1-\eta>0$, and since
$s_u/\sqrt m\to0$, $Z_u\to+\infty$ and $p_u\to0$. For a fixed family, the
smallest BY critical value $c_1=\alpha/(PL_P)$ is a positive constant. With
probability tending to one, every such $p_u$ is below $c_1$, and the step-up
procedure rejects every p-value below $c_1$.

(b) The statistic $Z_u$ equals $(\hat\mu_u-t_u)/(a_us_u/\sqrt m)$, where
$\hat\mu_u=a_u\bar A_u$ and $t_u=\max(\eta a_u,\tau_u)$, so its sign is that
of the margin $\hat\mu_u-t_u$. Suppose first that $r^\dagger_u\ne0$. Then
$a_u\to\|r^\dagger_u\|$, and
$\mu_u=\langle r_{D,u},c_u\rangle/\|r_{D,u}\|\to\|r^\dagger_u\|$, so the
margin converges to
$g_u=\|r^\dagger_u\|-\max(\eta\|r^\dagger_u\|,\tau^\dagger_u)$. If
$\|r^\dagger_u\|>\tau^\dagger_u$, then $g_u>0$ because also
$\|r^\dagger_u\|>\eta\|r^\dagger_u\|$; if $\|r^\dagger_u\|<\tau^\dagger_u$,
then $g_u\le\|r^\dagger_u\|-\tau^\dagger_u<0$. Suppose next that
$r^\dagger_u=0$. If $r_{D,u}=0$, then $p_u=1$. Otherwise
$|\mu_u|\le\|c_u\|\to0$ by the Cauchy--Schwarz inequality, whatever the
direction of $r_{D,u}$, while $t_u\ge\tau_u\to\tau^\dagger_u$, so the margin
tends to $-\tau^\dagger_u<0$. With vanishing standard errors, a positive
limit margin gives $p_u\to0$ and a negative one gives $p_u\to1$. The first
kind is rejected as in part (a); the second is not, because the largest
critical value $c_P=\alpha/L_P$ is below one.
\end{IEEEproof}

\noindent\textbf{A threshold on the share alone keeps every pattern that both
fits share, however faint; a floor on the reproduced strength keeps exactly
the pairs whose interaction exceeds it.} Part (a) asks only that the two
tables agree relative to their size, so it also covers tables that vanish. If
$r_{D,n}=c_n=\varepsilon_nB$ for a unit table $B$ and
$\varepsilon_n\downarrow0$, then $\theta_n=1$ for every $n$, and if the
context tables scale with $\varepsilon_n$, the whole ratio test does not
depend on $\varepsilon_n$. Shared faint patterns arise without any true
interaction; they are the shared bias of Proposition~\ref{prop:cross}. Let
$x_1,x_2$ be uniform on $\{-1,1\}$ and the true log-rate $ax_1+bx_2$. A
three-leaf tree that splits $x_2$ only where $x_1=1$ fits the residual table
$(b/2)x_1x_2$, and it does so on every sample. Correlated features add a
second source: when the data never visit some grid cells, a model is free to
extrapolate there, and the same learner tends to extrapolate the same way on
two samples. A model that converges to an additive function can thus hand
over a list that does not shrink.

The relative floor $\eta s_D$ removes pairs that are weak compared with the
rest of the family. It cannot remove patterns that shrink in every pair at
once, however: if every table is $\varepsilon_nB_u$, then $a_u=s_D=\varepsilon_n$
and the floor shrinks with them. This is the case of a model that has learned
no interaction, where every table is an artifact of the learner. The scale
floor $\kappa\sigma_D$ does not shrink with the tables, so part (b) covers
this case too.

No floor separates a true interaction from its proxies when features are
strongly correlated. If $x_3$ is a noisy copy of $x_1$, a model of $x_1x_2$
can route part of the interaction through the pair $(x_3,x_2)$, and the fitted
surface of that pair is then non-additive in every fit. LENS tests the
interactions of the model, so it reports the proxy when both fits route the
interaction the same way.

\section{Discussion}
\label{sec:discussion}

\textbf{The price of a split.} Each LENS model sees a third of the data, so
the list is the one that a third of the data can confirm. On freMTPL2 the
list grows from 6.6 to 12 pairs as the data per model grow from 20,000 to
184,000 policies, and most of the pairs kept with less data stay
(Section~\ref{sec:exp-real}), so a split mainly costs the weakest pairs. The
three parts need not be equal. The evaluation part only supplies contexts
for an average, and Section~\ref{sec:exp-sensitivity} shows that the
standard errors are rarely what decides, so it can be much smaller than the
two training parts. It must be fresh, however: the conditioning argument of
Section~\ref{sec:validity} needs contexts that are independent of both fits.
Rotating the roles of the parts would use the data more fully. Multi-split
methods combine p-values across random splits for a fixed set of
hypotheses~\cite{meinshausen2009pvalues}; here the hypotheses themselves
change with the rotation, because the discovery directions and boundaries
do, so keeping Theorem~\ref{thm:fdr} across rotations needs a new argument.
We use one split because its error control follows directly from the
conditioning.

\textbf{Setting the boundary.} The floor $\kappa\sigma_D$ is a size of
effect, so an analyst can replace it with a tolerance from the application.
In a motor tariff, a floor of 0.01 on the log claim rate says that the
reproduced interaction must move the claim rate by about 1\% over the grid
to be worth a term.

\textbf{Two fits of one learner.} LENS replicates across samples, not across
learners. Both fits come from the same algorithm, which is what lets them
share the systematic part of their errors (Proposition~\ref{prop:cross}),
and what the floor must then remove. Fitting the confirmation model with a
different learner would cancel some of that shared part. It would also change
the question from ``did this model learn a stable interaction'' to ``do two
kinds of model agree on it''. Our experiments hint at what this would do. On
Pair-corr, XGBoost routes part of the true interaction through the proxies
and the network almost does not; a cross-learner test would likely drop the
proxies, together with any true interaction that only one learner
represents. We have not evaluated this variant.

\textbf{What the averaged surface misses.} LENS tests the partial-dependence
surface, the average of the context surfaces. A three-way term
$x_jx_kx_\ell$ with $\mathbb Ex_\ell=0$ leaves every pair of
$\{j,k,\ell\}$ additive on average, and an interaction whose sign changes
across subgroups can cancel in the same way. The test inherits this from
partial dependence. Extrapolation to grid cells that the data never visit is
a related concern. A pattern that one fit invents there must replicate in the
other fit and exceed the floor. But a pattern that the learner extrapolates
the same way on both samples can pass both, which is one source of the shared
bias of Section~\ref{sec:boundary}. The same projection extends to triples,
with three-way tables in place of two-way ones and $p(p-1)(p-2)/6$
candidates, which keeps the two fits and multiplies the queries by the third
grid. We leave this extension to future work.

\section{Related Work}
\label{sec:related}

\textbf{False discoveries in data management.} The database community has long
asked how many results mined in bulk are real. Interactive exploration tests many
hypotheses implicitly and so invites false discoveries~\cite{zhao2017controlling}.
In association rule mining, the permutation correction is the most powerful but
also the most costly~\cite{liu2011controlling}, and FSR reduces the resampled
datasets that significant pattern mining needs~\cite{pellegrina2024efficient}.
CleanML adopts the Benjamini--Yekutieli procedure for the same reason as LENS: its
tests are dependent~\cite{li2021cleanml}. Closest to our setting, Slice Finder
tests the data slices on which a trained model
underperforms~\cite{chung2019slice,chung2020automated}. These methods test
patterns in the data or in the loss of a model. LENS asks the same question of
the interactions a model has learned.

\textbf{Ranking interactions.} Model-specific scores read one model family:
TreeSHAP for tree ensembles~\cite{lundberg2020local}, NID for network
weights~\cite{tsang2018detecting}, and Integrated Hessians for differentiable
networks~\cite{janizek2021explaining}. Model-agnostic scores include
$H^2$~\cite{friedman2008predictive}, built on functional
ANOVA~\cite{hooker2004discovering}, Archipelago~\cite{tsang2020does}, and
SHAP-IQ~\cite{fumagalli2023shapiq}. Other methods build interactions into the
model: ARM-Net learns cross features for structured data~\cite{cai2021armnet},
and FAST adds pairs to an additive model one at a time~\cite{lou2013accurate}.
None of them tests the pairs, so the length of the list is left to the user.
Others have also found $H^2$ unreliable~\cite{greenwell2018simple,
welchowski2024interaction,inglis2022visualizing}.

\textbf{Testing and replication.} Refit tests build a null distribution by
refitting the model~\cite{friedman2008predictive,lampa2014identification,
welchowski2022techniques}, but an empirical rank p-value from $B$ refits cannot fall below
$1/(B+1)$~\cite{phipson2010permutation}, so a correction over many pairs needs
many refits. DIAMOND controls the false discovery rate with model-X knockoffs,
which must reproduce the joint law of the features~\cite{chen2025diamond,
candes2018panning}; ARDA uses random features as a related baseline for single
features~\cite{chepurko2020arda}. iLOCO compares the loss after leaving out
features and pairs~\cite{little2025iloco}. For single features, sample-splitting
tests rest on the same fact as LENS: fitting noise in one model is independent
of a model trained on separate data~\cite{dai2024significance,tansey2022holdout}.
Statistics has formal notions of replicability across
studies~\cite{benjamini2008screening,bogomolov2013discovering,heller2014deciding}.
LENS applies replication to the interactions of a fitted model.

\section{Conclusion}
\label{sec:conclusion}

We presented LENS, a model-agnostic framework that selects feature interactions
through independent replication: it asks whether a second model, trained on
independent data, reproduces the interaction pattern of the first. It needs two
fits for any number of pairs, and one multiple-testing correction sets the
length of the list. Our theory shows why the unnormalized energy ranks
consistently where $H^2$ does not, and why consistent selection needs a floor on
the reproduced strength. The list is clean and finds most material interactions
on synthetic benchmarks, agrees with the actuarial literature and recovers most
of the gain of all pairs on insurance data, and keeps few pairs without raising
the loss on credit data. A natural next step is three-way interactions.

\section*{AI-Generated Content Acknowledgement}
Claude (Anthropic) assisted with language editing throughout the manuscript and with developing the experimental code. GPT Image (OpenAI) produced a layout draft for Fig.~\ref{fig:overview}, which we redrew as a vector figure. The authors reviewed and revised all content and take full responsibility for it.

\bibliographystyle{IEEEtran}
\bibliography{refs,extra}

\begin{thebibliography}{10}
\providecommand{\url}[1]{#1}
\csname url@samestyle\endcsname
\providecommand{\newblock}{\relax}
\providecommand{\bibinfo}[2]{#2}
\providecommand{\BIBentrySTDinterwordspacing}{\spaceskip=0pt\relax}
\providecommand{\BIBentryALTinterwordstretchfactor}{4}
\providecommand{\BIBentryALTinterwordspacing}{\spaceskip=\fontdimen2\font plus
\BIBentryALTinterwordstretchfactor\fontdimen3\font minus
  \fontdimen4\font\relax}
\providecommand{\BIBforeignlanguage}[2]{{%
\expandafter\ifx\csname l@#1\endcsname\relax
\typeout{** WARNING: IEEEtran.bst: No hyphenation pattern has been}%
\typeout{** loaded for the language `#1'. Using the pattern for}%
\typeout{** the default language instead.}%
\else
\language=\csname l@#1\endcsname
\fi
#2}}
\providecommand{\BIBdecl}{\relax}
\BIBdecl

\bibitem{cheng2016wide}
H.-T. Cheng, L.~Koc, J.~Harmsen, T.~Shaked, T.~Chandra, H.~Aradhye,
  G.~Anderson, G.~Corrado, W.~Chai, M.~Ispir, R.~Anil, Z.~Haque, L.~Hong,
  V.~Jain, X.~Liu, and H.~Shah, ``Wide \& deep learning for recommender
  systems,'' in \emph{Proceedings of the Workshop on Deep Learning for
  Recommender Systems}, 2016, pp. 7--10.

\bibitem{luo2019autocross}
Y.~Luo, M.~Wang, H.~Zhou, Q.~Yao, W.-W. Tu, Y.~Chen, W.~Dai, and Q.~Yang,
  ``{AutoCross}: Automatic feature crossing for tabular data in real-world
  applications,'' in \emph{Proceedings of the ACM SIGKDD International
  Conference on Knowledge Discovery and Data Mining}, 2019, pp. 1936--1945.

\bibitem{cordell2009detecting}
H.~J. Cordell, ``Detecting gene--gene interactions that underlie human
  diseases,'' \emph{Nature Reviews Genetics}, vol.~10, no.~6, pp. 392--404,
  2009.

\bibitem{schelldorfer2019nesting}
J.~Schelldorfer and M.~V. W{\"u}thrich, ``Nesting classical actuarial models
  into neural networks,'' SSRN 3320525, 2019.

\bibitem{lou2013accurate}
Y.~Lou, R.~Caruana, J.~Gehrke, and G.~Hooker, ``Accurate intelligible models
  with pairwise interactions,'' in \emph{Proceedings of the ACM SIGKDD
  International Conference on Knowledge Discovery and Data Mining}, 2013, pp.
  623--631.

\bibitem{havrylenko2024detection}
Y.~Havrylenko and J.~Heger, ``Detection of interacting variables for
  generalized linear models via neural networks,'' \emph{European Actuarial
  Journal}, vol.~14, no.~2, pp. 551--580, 2024.

\bibitem{lindholm2024blackbox}
M.~Lindholm and J.~Palmquist, ``Black-box guided generalised linear model
  building with non-life pricing applications,'' \emph{Annals of Actuarial
  Science}, vol.~18, no.~3, pp. 675--691, 2024.

\bibitem{lundberg2020local}
S.~M. Lundberg, G.~Erion, H.~Chen, A.~DeGrave, J.~M. Prutkin, B.~Nair, R.~Katz,
  J.~Himmelfarb, N.~Bansal, and S.-I. Lee, ``From local explanations to global
  understanding with explainable {AI} for trees,'' \emph{Nature Machine
  Intelligence}, vol.~2, no.~1, pp. 56--67, 2020.

\bibitem{tsang2018detecting}
M.~Tsang, D.~Cheng, and Y.~Liu, ``Detecting statistical interactions from
  neural network weights,'' in \emph{Proceedings of the International
  Conference on Learning Representations}, 2018.

\bibitem{janizek2021explaining}
J.~D. Janizek, P.~Sturmfels, and S.-I. Lee, ``Explaining explanations:
  Axiomatic feature interactions for deep networks,'' \emph{Journal of Machine
  Learning Research}, vol.~22, no. 104, pp. 1--54, 2021.

\bibitem{friedman2008predictive}
J.~H. Friedman and B.~E. Popescu, ``Predictive learning via rule ensembles,''
  \emph{The Annals of Applied Statistics}, vol.~2, no.~3, pp. 916--954, 2008.

\bibitem{henckaerts2022stakes}
R.~Henckaerts, K.~Antonio, and M.-P. C{\^o}t{\'e}, ``When stakes are high:
  Balancing accuracy and transparency with model-agnostic interpretable
  data-driven surrogates,'' \emph{Expert Systems with Applications}, vol. 202,
  2022, {Art}. no. 117230.

\bibitem{lampa2014identification}
E.~Lampa, L.~Lind, P.~M. Lind, and A.~Bornefalk-Hermansson, ``The
  identification of complex interactions in epidemiology and toxicology: A
  simulation study of boosted regression trees,'' \emph{Environmental Health},
  vol.~13, 2014, {Art}. no. 57.

\bibitem{welchowski2022techniques}
T.~Welchowski, K.~O. Maloney, R.~Mitchell, and M.~Schmid, ``Techniques to
  improve ecological interpretability of black-box machine learning models,''
  \emph{Journal of Agricultural, Biological and Environmental Statistics},
  vol.~27, no.~1, pp. 175--197, 2022.

\bibitem{little2025iloco}
C.~Little, L.~Zheng, and G.~Allen, ``{iLOCO}: Distribution-free inference for
  feature interactions,'' \emph{arXiv preprint arXiv:2502.06661}, 2025.

\bibitem{chen2025diamond}
W.~Chen, Y.~Jiang, W.~S. Noble, and Y.~Y. Lu, ``Error-controlled non-additive
  interaction discovery in machine learning models,'' \emph{Nature Machine
  Intelligence}, vol.~7, no.~9, pp. 1541--1554, 2025.

\bibitem{benjamini2001control}
Y.~Benjamini and D.~Yekutieli, ``The control of the false discovery rate in
  multiple testing under dependency,'' \emph{The Annals of Statistics},
  vol.~29, no.~4, pp. 1165--1188, 2001.

\bibitem{inglis2022visualizing}
A.~Inglis, A.~Parnell, and C.~B. Hurley, ``Visualizing variable importance and
  variable interaction effects in machine learning models,'' \emph{Journal of
  Computational and Graphical Statistics}, vol.~31, no.~3, pp. 766--778, 2022.

\bibitem{muschalik2024beyond}
M.~Muschalik, F.~Fumagalli, B.~Hammer, and E.~H{\"u}llermeier, ``Beyond
  {TreeSHAP}: Efficient computation of any-order {Shapley} interactions for
  tree ensembles,'' in \emph{Proceedings of the AAAI Conference on Artificial
  Intelligence}, vol.~38, 2024, pp. 14\,388--14\,396.

\bibitem{tsang2020does}
M.~Tsang, S.~Rambhatla, and Y.~Liu, ``How does this interaction affect me?
  {Interpretable} attribution for feature interactions,'' in \emph{Advances in
  Neural Information Processing Systems}, vol.~33, 2020, pp. 6147--6159.

\bibitem{fumagalli2023shapiq}
F.~Fumagalli, M.~Muschalik, P.~Kolpaczki, E.~H{\"u}llermeier, and B.~Hammer,
  ``{SHAP-IQ}: Unified approximation of any-order {Shapley} interactions,'' in
  \emph{Advances in Neural Information Processing Systems}, vol.~36, 2023, pp.
  11\,515--11\,551.

\bibitem{tsai2023faith}
C.-P. Tsai, C.-K. Yeh, and P.~Ravikumar, ``{Faith-Shap}: The faithful {Shapley}
  interaction index,'' \emph{Journal of Machine Learning Research}, vol.~24,
  no.~94, pp. 1--42, 2023.

\bibitem{mayer2023hstats}
M.~Mayer, ``hstats: Interaction statistics,'' R package, CRAN, 2023.

\bibitem{basu2018iterative}
S.~Basu, K.~Kumbier, J.~B. Brown, and B.~Yu, ``Iterative random forests to
  discover predictive and stable high-order interactions,'' \emph{Proceedings
  of the National Academy of Sciences}, vol. 115, no.~8, pp. 1943--1948, 2018.

\bibitem{meinshausen2009pvalues}
N.~Meinshausen, L.~Meier, and P.~B{\"u}hlmann, ``{$p$}-values for
  high-dimensional regression,'' \emph{Journal of the American Statistical
  Association}, vol. 104, no. 488, pp. 1671--1681, 2009.

\bibitem{zhao2017controlling}
Z.~Zhao, L.~De~Stefani, E.~Zgraggen, C.~Binnig, E.~Upfal, and T.~Kraska,
  ``Controlling false discoveries during interactive data exploration,'' in
  \emph{Proceedings of the ACM SIGMOD International Conference on Management of
  Data}, 2017, pp. 527--540.

\bibitem{liu2011controlling}
G.~Liu, H.~Zhang, and L.~Wong, ``Controlling false positives in association
  rule mining,'' \emph{Proceedings of the VLDB Endowment}, vol.~5, no.~2, pp.
  145--156, 2011.

\bibitem{pellegrina2024efficient}
L.~Pellegrina and F.~Vandin, ``Efficient discovery of significant patterns with
  few-shot resampling,'' \emph{Proceedings of the VLDB Endowment}, vol.~17,
  no.~10, pp. 2668--2680, 2024.

\bibitem{li2021cleanml}
P.~Li, S.~X. Rao, J.~L. Blase, Y.~Zhang, X.~Chu, and C.~Zhang, ``{CleanML}: A
  study for evaluating the impact of data cleaning on {ML} classification
  tasks,'' in \emph{Proceedings of the IEEE International Conference on Data
  Engineering}, 2021, pp. 13--24.

\bibitem{chung2019slice}
Y.~Chung, T.~Kraska, N.~Polyzotis, K.~H. Tae, and S.~E. Whang, ``{Slice
  Finder}: Automated data slicing for model validation,'' in \emph{Proceedings
  of the IEEE International Conference on Data Engineering}, 2019, pp.
  1550--1553.

\bibitem{chung2020automated}
Y.~Chung, T.~Kraska, N.~Polyzotis, K.~H. Tae, and S.~E. Whang, ``Automated data
  slicing for model validation: A big data--{AI} integration approach,''
  \emph{IEEE Transactions on Knowledge and Data Engineering}, vol.~32, no.~12,
  pp. 2284--2296, 2020.

\bibitem{hooker2004discovering}
G.~Hooker, ``Discovering additive structure in black box functions,'' in
  \emph{Proceedings of the ACM SIGKDD International Conference on Knowledge
  Discovery and Data Mining}, 2004, pp. 575--580.

\bibitem{cai2021armnet}
S.~Cai, K.~Zheng, G.~Chen, H.~V. Jagadish, B.~C. Ooi, and M.~Zhang,
  ``{ARM-Net}: Adaptive relation modeling network for structured data,'' in
  \emph{Proceedings of the ACM SIGMOD International Conference on Management of
  Data}, 2021, pp. 207--220.

\bibitem{greenwell2018simple}
B.~M. Greenwell, B.~C. Boehmke, and A.~J. McCarthy, ``A simple and effective
  model-based variable importance measure,'' \emph{arXiv preprint
  arXiv:1805.04755}, 2018.

\bibitem{welchowski2024interaction}
T.~Welchowski and D.~Edelmann, ``Interaction difference hypothesis test for
  prediction models,'' \emph{Machine Learning and Knowledge Extraction},
  vol.~6, no.~2, pp. 1298--1322, 2024.

\bibitem{phipson2010permutation}
B.~Phipson and G.~K. Smyth, ``Permutation {$p$}-values should never be zero:
  Calculating exact {$p$}-values when permutations are randomly drawn,''
  \emph{Statistical Applications in Genetics and Molecular Biology}, vol.~9,
  no.~1, 2010, {Art}. no. 39.

\bibitem{candes2018panning}
E.~Cand{\`e}s, Y.~Fan, L.~Janson, and J.~Lv, ``Panning for gold: `{Model-X}'
  knockoffs for high dimensional controlled variable selection,'' \emph{Journal
  of the Royal Statistical Society: Series B}, vol.~80, no.~3, pp. 551--577,
  2018.

\bibitem{chepurko2020arda}
N.~Chepurko, R.~Marcus, E.~Zgraggen, R.~C. Fernandez, T.~Kraska, and D.~Karger,
  ``{ARDA}: Automatic relational data augmentation for machine learning,''
  \emph{Proceedings of the VLDB Endowment}, vol.~13, no.~9, pp. 1373--1387,
  2020.

\bibitem{dai2024significance}
B.~Dai, X.~Shen, and W.~Pan, ``Significance tests of feature relevance for a
  black-box learner,'' \emph{IEEE Transactions on Neural Networks and Learning
  Systems}, vol.~35, no.~2, pp. 1898--1911, 2024.

\bibitem{tansey2022holdout}
W.~Tansey, V.~Veitch, H.~Zhang, R.~Rabadan, and D.~M. Blei, ``The holdout
  randomization test for feature selection in black box models,'' \emph{Journal
  of Computational and Graphical Statistics}, vol.~31, no.~1, pp. 151--162,
  2022.

\bibitem{benjamini2008screening}
Y.~Benjamini and R.~Heller, ``Screening for partial conjunction hypotheses,''
  \emph{Biometrics}, vol.~64, no.~4, pp. 1215--1222, 2008.

\bibitem{bogomolov2013discovering}
M.~Bogomolov and R.~Heller, ``Discovering findings that replicate from a
  primary study of high dimension to a follow-up study,'' \emph{Journal of the
  American Statistical Association}, vol. 108, no. 504, pp. 1480--1492, 2013.

\bibitem{heller2014deciding}
R.~Heller, M.~Bogomolov, and Y.~Benjamini, ``Deciding whether follow-up studies
  have replicated findings in a preliminary large-scale omics study,''
  \emph{Proceedings of the National Academy of Sciences}, vol. 111, no.~46, pp.
  16\,262--16\,267, 2014.

\end{thebibliography}

\clearpage
\onecolumn
\appendices

\section{Complete Synthetic Results}
\label{app:synthetic}

Tables~\ref{tab:app-sparse_pairs}--\ref{tab:app-additive_corr} give every
design at every number of features, for every method and every variant of its
decision rule that we ran. Table~\ref{tab:lists} in the body reports the
default rule of each method at $p=50$. LENS appears with its default boundary
($\kappa=0.02$), with a smaller scale floor ($\kappa=0.01$), without the scale
floor ($\kappa=0$, the boundary $\max(\eta a_u,\eta s_D)$), and with the purely
relative boundary $b_u=\eta$ of Proposition~\ref{prop:floor}(a). DIAMOND
appears at both levels of its knockoff filter, with its own XGBoost, with the
XGBoost settings of LENS, and with its MLP. $H^2$ appears with the rank
p-values of the refit test, which keep no pair, and with the normal
approximation used in the body. Tables~\ref{tab:app-f-fdp}--\ref{tab:app-f-size}
break F1--F10 down by function.

\begin{table}[H]\centering
\caption{Sparse design: list size, FDP, and power, means over 20 seeds, for $p=10$, 30, and 50 (45, 435, and 1,225 candidate pairs).}
\label{tab:app-sparse_pairs}\footnotesize\setlength{\tabcolsep}{4pt}
\begin{tabular}{llccccccccc}
\toprule
 & & \multicolumn{3}{c}{$p=10$} & \multicolumn{3}{c}{$p=30$} & \multicolumn{3}{c}{$p=50$} \\
\cmidrule(lr){3-5}\cmidrule(lr){6-8}\cmidrule(lr){9-11}
Method & Model & size & FDP & power & size & FDP & power & size & FDP & power \\
\midrule
LENS (default, $\kappa=0.02$) & XGBoost & 3.0 & 0.01 & 1.00 & 3.0 & 0.01 & 1.00 & 3.0 & 0.01 & 1.00 \\
LENS, $\kappa=0.01$ & XGBoost & 3.0 & 0.01 & 1.00 & 3.2 & 0.06 & 1.00 & 3.4 & 0.09 & 1.00 \\
LENS, $\kappa=0$ & XGBoost & 3.0 & 0.01 & 1.00 & 3.5 & 0.12 & 1.00 & 7.2 & 0.54 & 1.00 \\
LENS, $b_u=\eta$ & XGBoost & 8.3 & 0.61 & 1.00 & 79.0 & 0.96 & 1.00 & 233.8 & 0.99 & 1.00 \\
LENS (default, $\kappa=0.02$) & MLP & 3.0 & 0.00 & 1.00 & 3.0 & 0.00 & 1.00 & 3.0 & 0.00 & 1.00 \\
LENS, $\kappa=0.01$ & MLP & 3.0 & 0.00 & 1.00 & 3.0 & 0.00 & 1.00 & 3.0 & 0.00 & 1.00 \\
LENS, $\kappa=0$ & MLP & 3.0 & 0.00 & 1.00 & 3.0 & 0.00 & 1.00 & 3.9 & 0.20 & 1.00 \\
DIAMOND, $q=0.2$ & XGBoost & 2.8 & 0.03 & 0.90 & 3.5 & 0.16 & 0.82 & 4.0 & 0.18 & 0.93 \\
DIAMOND, $q=0.05$ & XGBoost & 2.8 & 0.03 & 0.90 & 3.1 & 0.14 & 0.82 & 3.9 & 0.17 & 0.93 \\
DIAMOND, $q=0.2$ & XGBoost$^\dagger$ & 2.9 & 0.04 & 0.90 & 3.6 & 0.19 & 0.82 & 4.0 & 0.17 & 0.93 \\
DIAMOND, $q=0.2$ & MLP & 2.8 & 0.01 & 0.90 & 3.4 & 0.16 & 0.82 & 3.6 & 0.14 & 0.93 \\
DIAMOND, $q=0.05$ & MLP & 2.8 & 0.01 & 0.90 & 3.0 & 0.14 & 0.82 & 3.6 & 0.14 & 0.93 \\
$H^2$, normal refit null & XGBoost & 3.5 & 0.11 & 1.00 & 36.5 & 0.91 & 1.00 & 73.5 & 0.96 & 1.00 \\
$H^2$, rank p-values & XGBoost & 0.0 & 0.00 & 0.00 & 0.0 & 0.00 & 0.00 & 0.0 & 0.00 & 0.00 \\
iLOCO-Split, $\alpha=0.05$ & XGBoost & 15.2 & 0.76 & 1.00 & 118.5 & 0.92 & 1.00 & 399.1 & 0.98 & 1.00 \\
iLOCO-Split, $\alpha=0.2$ & XGBoost & 19.2 & 0.82 & 1.00 & 136.9 & 0.94 & 1.00 & 437.8 & 0.98 & 1.00 \\
iLOCO-MP, $\alpha=0.05$ & trees & 13.4 & 0.77 & 1.00 & 205.8 & 0.99 & 1.00 & 487.6 & 0.99 & 1.00 \\
FAST & EBM & 16.1 & 0.78 & 1.00 & 19.0 & 0.83 & 1.00 & 20.0 & 0.85 & 1.00 \\
\bottomrule\end{tabular}
\tabnote{$^\dagger$ DIAMOND with the XGBoost settings of LENS.}
\end{table}

\begin{table}[H]\centering
\caption{Pair-corr design: list size, FDP, and power, means over 20 seeds, for $p=10$, 30, and 50 (45, 435, and 1,225 candidate pairs).}
\label{tab:app-pair_corr}\footnotesize\setlength{\tabcolsep}{4pt}
\begin{tabular}{llccccccccc}
\toprule
 & & \multicolumn{3}{c}{$p=10$} & \multicolumn{3}{c}{$p=30$} & \multicolumn{3}{c}{$p=50$} \\
\cmidrule(lr){3-5}\cmidrule(lr){6-8}\cmidrule(lr){9-11}
Method & Model & size & FDP & power & size & FDP & power & size & FDP & power \\
\midrule
LENS (default, $\kappa=0.02$) & XGBoost & 3.0 & 0.67 & 1.00 & 3.0 & 0.67 & 1.00 & 3.0 & 0.67 & 1.00 \\
LENS, $\kappa=0.01$ & XGBoost & 3.0 & 0.67 & 1.00 & 3.2 & 0.69 & 1.00 & 3.4 & 0.70 & 1.00 \\
LENS, $\kappa=0$ & XGBoost & 3.0 & 0.67 & 1.00 & 3.5 & 0.71 & 1.00 & 4.9 & 0.78 & 1.00 \\
LENS, $b_u=\eta$ & XGBoost & 8.6 & 0.88 & 1.00 & 76.8 & 0.99 & 1.00 & 240.8 & 1.00 & 1.00 \\
LENS (default, $\kappa=0.02$) & MLP & 1.1 & 0.03 & 1.00 & 1.1 & 0.05 & 1.00 & 1.1 & 0.03 & 1.00 \\
LENS, $\kappa=0.01$ & MLP & 1.1 & 0.03 & 1.00 & 2.1 & 0.40 & 1.00 & 2.4 & 0.36 & 1.00 \\
LENS, $\kappa=0$ & MLP & 1.1 & 0.03 & 1.00 & 2.8 & 0.58 & 1.00 & 12.3 & 0.90 & 1.00 \\
DIAMOND, $q=0.2$ & XGBoost & 0.5 & 0.11 & 0.20 & 1.3 & 0.24 & 0.35 & 2.0 & 0.35 & 0.55 \\
DIAMOND, $q=0.05$ & XGBoost & 0.5 & 0.11 & 0.20 & 1.2 & 0.23 & 0.35 & 1.8 & 0.35 & 0.55 \\
DIAMOND, $q=0.2$ & XGBoost$^\dagger$ & 0.5 & 0.12 & 0.20 & 1.7 & 0.30 & 0.40 & 1.6 & 0.31 & 0.50 \\
DIAMOND, $q=0.2$ & MLP & 0.6 & 0.13 & 0.25 & 1.4 & 0.27 & 0.35 & 0.6 & 0.12 & 0.20 \\
DIAMOND, $q=0.05$ & MLP & 0.6 & 0.13 & 0.25 & 1.2 & 0.27 & 0.35 & 0.6 & 0.12 & 0.20 \\
$H^2$, normal refit null & XGBoost & 9.1 & 0.88 & 1.00 & 54.3 & 0.98 & 1.00 & 87.4 & 0.99 & 1.00 \\
$H^2$, rank p-values & XGBoost & 0.0 & 0.00 & 0.00 & 0.0 & 0.00 & 0.00 & 0.0 & 0.00 & 0.00 \\
iLOCO-Split, $\alpha=0.05$ & XGBoost & 5.2 & 0.60 & 1.00 & 91.2 & 0.97 & 1.00 & 181.6 & 0.99 & 0.85 \\
iLOCO-Split, $\alpha=0.2$ & XGBoost & 6.5 & 0.65 & 1.00 & 118.2 & 0.98 & 1.00 & 226.8 & 0.99 & 0.90 \\
iLOCO-MP, $\alpha=0.05$ & trees & 19.0 & 0.95 & 1.00 & 256.9 & 1.00 & 1.00 & 588.6 & 1.00 & 1.00 \\
FAST & EBM & 9.6 & 0.71 & 1.00 & 9.3 & 0.48 & 1.00 & 8.9 & 0.48 & 1.00 \\
\bottomrule\end{tabular}
\tabnote{$^\dagger$ DIAMOND with the XGBoost settings of LENS.}
\end{table}

\begin{table}[H]\centering
\caption{Additive design: list size and the share of runs that return any pair (every pair is false), means over 20 seeds, for $p=10$, 30, and 50 (45, 435, and 1,225 candidate pairs).}
\label{tab:app-additive}\footnotesize\setlength{\tabcolsep}{4pt}
\begin{tabular}{llcccccc}
\toprule
 & & \multicolumn{2}{c}{$p=10$} & \multicolumn{2}{c}{$p=30$} & \multicolumn{2}{c}{$p=50$} \\
\cmidrule(lr){3-4}\cmidrule(lr){5-6}\cmidrule(lr){7-8}
Method & Model & size & any & size & any & size & any \\
\midrule
LENS (default, $\kappa=0.02$) & XGBoost & 0.1 & 0.05 & 0.1 & 0.05 & 0.1 & 0.10 \\
LENS, $\kappa=0.01$ & XGBoost & 0.1 & 0.15 & 0.2 & 0.20 & 0.2 & 0.25 \\
LENS, $\kappa=0$ & XGBoost & 0.9 & 0.65 & 7.5 & 1.00 & 17.2 & 1.00 \\
LENS, $b_u=\eta$ & XGBoost & 8.3 & 1.00 & 76.5 & 1.00 & 219.6 & 1.00 \\
LENS (default, $\kappa=0.02$) & MLP & 0.0 & 0.00 & 0.0 & 0.00 & 0.0 & 0.00 \\
LENS, $\kappa=0.01$ & MLP & 0.1 & 0.05 & 0.1 & 0.05 & 0.2 & 0.25 \\
LENS, $\kappa=0$ & MLP & 16.1 & 1.00 & 145.8 & 1.00 & 403.2 & 1.00 \\
DIAMOND, $q=0.2$ & XGBoost & 1.1 & 0.40 & 2.2 & 0.85 & 2.6 & 0.90 \\
DIAMOND, $q=0.05$ & XGBoost & 1.1 & 0.40 & 2.2 & 0.85 & 2.6 & 0.90 \\
DIAMOND, $q=0.2$ & XGBoost$^\dagger$ & 1.3 & 0.50 & 2.2 & 0.80 & 2.5 & 0.85 \\
DIAMOND, $q=0.2$ & MLP & 0.7 & 0.30 & 0.4 & 0.15 & 1.8 & 0.65 \\
DIAMOND, $q=0.05$ & MLP & 0.7 & 0.30 & 0.4 & 0.15 & 1.8 & 0.65 \\
$H^2$, normal refit null & XGBoost & 0.6 & 0.35 & 8.0 & 1.00 & 25.1 & 1.00 \\
$H^2$, rank p-values & XGBoost & 0.0 & 0.00 & 0.0 & 0.00 & 0.0 & 0.00 \\
iLOCO-Split, $\alpha=0.05$ & XGBoost & 1.4 & 0.35 & 1.2 & 0.20 & 9.3 & 0.45 \\
iLOCO-Split, $\alpha=0.2$ & XGBoost & 3.0 & 0.60 & 5.8 & 0.40 & 21.7 & 0.50 \\
iLOCO-MP, $\alpha=0.05$ & trees & 21.0 & 1.00 & 264.4 & 1.00 & 609.5 & 1.00 \\
FAST & EBM & 5.0 & 0.50 & 1.1 & 0.20 & 1.6 & 0.30 \\
\bottomrule\end{tabular}
\tabnote{$^\dagger$ DIAMOND with the XGBoost settings of LENS.}
\end{table}

\begin{table}[H]\centering
\caption{Additive-corr design: list size and the share of runs that return any pair (every pair is false), means over 20 seeds, for $p=10$, 30, and 50 (45, 435, and 1,225 candidate pairs).}
\label{tab:app-additive_corr}\footnotesize\setlength{\tabcolsep}{4pt}
\begin{tabular}{llcccccc}
\toprule
 & & \multicolumn{2}{c}{$p=10$} & \multicolumn{2}{c}{$p=30$} & \multicolumn{2}{c}{$p=50$} \\
\cmidrule(lr){3-4}\cmidrule(lr){5-6}\cmidrule(lr){7-8}
Method & Model & size & any & size & any & size & any \\
\midrule
LENS (default, $\kappa=0.02$) & XGBoost & 0.1 & 0.05 & 0.1 & 0.05 & 0.1 & 0.05 \\
LENS, $\kappa=0.01$ & XGBoost & 0.5 & 0.40 & 0.6 & 0.45 & 0.5 & 0.50 \\
LENS, $\kappa=0$ & XGBoost & 1.9 & 0.95 & 10.1 & 1.00 & 20.2 & 1.00 \\
LENS, $b_u=\eta$ & XGBoost & 9.4 & 1.00 & 92.4 & 1.00 & 228.4 & 1.00 \\
LENS (default, $\kappa=0.02$) & MLP & 0.2 & 0.20 & 0.1 & 0.10 & 0.1 & 0.15 \\
LENS, $\kappa=0.01$ & MLP & 1.4 & 0.90 & 1.1 & 0.75 & 1.1 & 0.70 \\
LENS, $\kappa=0$ & MLP & 13.4 & 1.00 & 137.4 & 1.00 & 377.7 & 1.00 \\
DIAMOND, $q=0.2$ & XGBoost & 1.4 & 0.40 & 2.3 & 0.80 & 3.0 & 0.90 \\
DIAMOND, $q=0.05$ & XGBoost & 1.4 & 0.40 & 2.3 & 0.80 & 2.9 & 0.90 \\
DIAMOND, $q=0.2$ & XGBoost$^\dagger$ & 1.2 & 0.35 & 2.4 & 0.80 & 3.0 & 0.90 \\
DIAMOND, $q=0.2$ & MLP & 0.6 & 0.15 & 1.1 & 0.45 & 3.1 & 0.90 \\
DIAMOND, $q=0.05$ & MLP & 0.5 & 0.15 & 1.1 & 0.45 & 3.1 & 0.90 \\
$H^2$, normal refit null & XGBoost & 0.4 & 0.20 & 9.7 & 0.95 & 23.6 & 1.00 \\
$H^2$, rank p-values & XGBoost & 0.0 & 0.00 & 0.0 & 0.00 & 0.0 & 0.00 \\
iLOCO-Split, $\alpha=0.05$ & XGBoost & 0.3 & 0.25 & 2.1 & 0.15 & 25.4 & 0.15 \\
iLOCO-Split, $\alpha=0.2$ & XGBoost & 1.6 & 0.40 & 4.6 & 0.30 & 33.1 & 0.30 \\
iLOCO-MP, $\alpha=0.05$ & trees & 22.2 & 1.00 & 247.7 & 1.00 & 599.3 & 1.00 \\
FAST & EBM & 4.3 & 0.60 & 2.2 & 0.50 & 2.5 & 0.60 \\
\bottomrule\end{tabular}
\tabnote{$^\dagger$ DIAMOND with the XGBoost settings of LENS.}
\end{table}

\begin{table}[H]\centering
\caption{F1--F10 ($p=30$): FDP per function, means over 20 seeds.}
\label{tab:app-f-fdp}\footnotesize\setlength{\tabcolsep}{3.5pt}
\begin{tabular}{llccccccccccc}
\toprule
Method & Model & F1 & F2 & F3 & F4 & F5 & F6 & F7 & F8 & F9 & F10 & mean \\
\midrule
LENS (default, $\kappa=0.02$) & XGBoost & 0.00 & 0.00 & 0.00 & 0.00 & 0.00 & 0.00 & 0.00 & 0.00 & 0.00 & 0.00 & 0.00 \\
LENS (default, $\kappa=0.02$) & MLP & 0.00 & 0.00 & 0.00 & 0.00 & 0.00 & 0.00 & 0.00 & 0.00 & 0.00 & 0.00 & 0.00 \\
DIAMOND, $q=0.2$ & XGBoost & 0.11 & 0.13 & 0.11 & 0.09 & 0.11 & 0.03 & 0.05 & 0.03 & 0.02 & 0.07 & 0.07 \\
DIAMOND, $q=0.2$ & MLP & 0.16 & 0.18 & 0.14 & 0.13 & 0.16 & 0.05 & 0.03 & 0.02 & 0.02 & 0.07 & 0.10 \\
$H^2$, normal refit null & XGBoost & 0.76 & 0.78 & 0.82 & 0.77 & 0.70 & 0.83 & 0.31 & 0.70 & 0.72 & 0.83 & 0.72 \\
iLOCO-Split, $\alpha=0.05$ & XGBoost & 0.72 & 0.64 & 0.76 & 0.76 & 0.49 & 0.70 & 0.63 & 0.61 & 0.62 & 0.77 & 0.67 \\
iLOCO-MP, $\alpha=0.05$ & trees & 0.98 & 0.98 & 0.99 & 0.99 & 0.99 & 0.98 & 1.00 & 0.99 & 0.98 & 0.99 & 0.99 \\
FAST & EBM & 0.51 & 0.54 & 0.46 & 0.47 & 0.57 & 0.68 & 0.35 & 0.38 & 0.40 & 0.70 & 0.51 \\
\bottomrule\end{tabular}\end{table}

\begin{table}[H]\centering
\caption{F1--F10 ($p=30$): power per function, means over 20 seeds.}
\label{tab:app-f-power}\footnotesize\setlength{\tabcolsep}{3.5pt}
\begin{tabular}{llccccccccccc}
\toprule
Method & Model & F1 & F2 & F3 & F4 & F5 & F6 & F7 & F8 & F9 & F10 & mean \\
\midrule
LENS (default, $\kappa=0.02$) & XGBoost & 0.54 & 0.55 & 1.00 & 1.00 & 0.88 & 0.25 & 0.21 & 0.67 & 0.13 & 1.00 & 0.62 \\
LENS (default, $\kappa=0.02$) & MLP & 0.64 & 0.64 & 1.00 & 1.00 & 0.88 & 0.25 & 0.21 & 0.75 & 0.27 & 1.00 & 0.66 \\
DIAMOND, $q=0.2$ & XGBoost & 0.29 & 0.29 & 0.40 & 0.35 & 0.76 & 0.12 & 0.14 & 0.26 & 0.06 & 0.47 & 0.31 \\
DIAMOND, $q=0.2$ & MLP & 0.29 & 0.28 & 0.16 & 0.20 & 0.75 & 0.12 & 0.15 & 0.28 & 0.10 & 0.59 & 0.29 \\
$H^2$, normal refit null & XGBoost & 0.94 & 0.93 & 1.00 & 1.00 & 0.88 & 1.00 & 0.84 & 1.00 & 1.00 & 1.00 & 0.96 \\
iLOCO-Split, $\alpha=0.05$ & XGBoost & 0.57 & 0.55 & 0.94 & 0.95 & 0.76 & 0.81 & 0.02 & 0.19 & 0.56 & 0.64 & 0.60 \\
iLOCO-MP, $\alpha=0.05$ & trees & 0.40 & 0.38 & 0.17 & 0.15 & 0.46 & 0.62 & 0.04 & 0.15 & 0.38 & 0.27 & 0.30 \\
FAST & EBM & 0.81 & 0.80 & 1.00 & 1.00 & 0.88 & 0.80 & 0.93 & 1.00 & 0.80 & 1.00 & 0.90 \\
\bottomrule\end{tabular}\end{table}

\begin{table}[H]\centering
\caption{F1--F10 ($p=30$): list size per function, means over 20 seeds.}
\label{tab:app-f-size}\footnotesize\setlength{\tabcolsep}{3.5pt}
\begin{tabular}{llccccccccccc}
\toprule
Method & Model & F1 & F2 & F3 & F4 & F5 & F6 & F7 & F8 & F9 & F10 & mean \\
\midrule
LENS (default, $\kappa=0.02$) & XGBoost & 6.0 & 6.0 & 9.0 & 10.0 & 7.0 & 2.0 & 3.0 & 8.0 & 2.0 & 6.0 & 5.9 \\
LENS (default, $\kappa=0.02$) & MLP & 7.0 & 7.0 & 9.0 & 10.0 & 7.0 & 2.0 & 3.0 & 9.0 & 4.0 & 6.0 & 6.4 \\
DIAMOND, $q=0.2$ & XGBoost & 3.6 & 3.7 & 4.5 & 4.2 & 7.0 & 1.0 & 2.1 & 3.3 & 0.9 & 3.2 & 3.4 \\
DIAMOND, $q=0.2$ & MLP & 4.0 & 3.9 & 2.4 & 3.1 & 7.3 & 1.1 & 2.2 & 3.5 & 1.6 & 4.0 & 3.3 \\
$H^2$, normal refit null & XGBoost & 43.2 & 46.5 & 50.3 & 45.2 & 24.5 & 48.6 & 17.5 & 41.1 & 56.0 & 36.2 & 40.9 \\
iLOCO-Split, $\alpha=0.05$ & XGBoost & 34.4 & 35.8 & 67.7 & 52.5 & 19.4 & 39.1 & 3.4 & 18.1 & 27.7 & 38.1 & 33.6 \\
iLOCO-MP, $\alpha=0.05$ & trees & 253.2 & 251.7 & 220.3 & 225.2 & 249.9 & 272.4 & 226.3 & 255.1 & 270.4 & 260.8 & 248.5 \\
FAST & EBM & 18.8 & 19.6 & 17.6 & 19.2 & 17.4 & 20.0 & 20.0 & 19.6 & 20.0 & 20.0 & 19.2 \\
\bottomrule\end{tabular}\end{table}

\section{Which Part of LENS Does the Work}
\label{app:ablation}

LENS combines four ingredients: a ranking by the energy $a_u^2$, a second
model, a comparison of the two fits along the discovery pattern, and a test.
Each of the following rules removes one of them and keeps the boundary of
Eq.~\eqref{eq:null}.
\begin{itemize}
\item \emph{Top-$|\widehat G|$ by energy} removes the decision: it keeps the
  $|\widehat G|$ pairs of highest energy, as many as LENS keeps. It isolates
  which pairs the test chooses, at a fixed length.
\item \emph{$\hat f_D$ only} removes the second model: it keeps every pair
  whose discovery strength $a_u$ exceeds the floor
  $\max(\eta s_D,\kappa\sigma_D)$.
\item \emph{Both fits above the floor, no comparison} uses the second model
  but not the comparison: it keeps a pair when both $a_u$ and the strength
  $\|r_{C,u}\|$ of the confirmation table exceed the same floor, whatever the
  direction of the two tables.
\item \emph{$\hat\mu_u$ above the boundary, no test} keeps the comparison and
  drops the test: it keeps a pair when the reproduced strength
  $\hat\mu_u=a_u\bar A_u$ exceeds $\max(\eta a_u,\eta s_D,\kappa\sigma_D)$.
\end{itemize}
All four are computed from the same two fits as LENS, so the comparison
differs only in the rule.

\begin{table}[H]\centering
\caption{Ablations on synthetic data with XGBoost, means over 20 seeds. Each rule removes one ingredient of LENS and keeps the boundary of Eq.~\eqref{eq:null}. For the additive designs, the share of runs that return any pair (every pair is false); otherwise list size, FDP, and power.}
\label{tab:app-ablation}\footnotesize\setlength{\tabcolsep}{3.2pt}
\begin{tabular}{lcc ccc ccc ccc}
\toprule
 & Additive & Additive-corr & \multicolumn{3}{c}{Sparse} & \multicolumn{3}{c}{Pair-corr} & \multicolumn{3}{c}{F1--F10 ($p=30$)} \\
\cmidrule(lr){4-6}\cmidrule(lr){7-9}\cmidrule(lr){10-12}
Rule & any & any & size & FDP & power & size & FDP & power & size & FDP & power \\
\midrule
\multicolumn{12}{l}{\emph{$p=10$}} \\
LENS (default) & 0.05 & 0.05 & 3.0 & 0.01 & 1.00 & 3.0 & 0.67 & 1.00 &  &  &  \\
$\hat\mu_u$ above the boundary, no test & 0.05 & 0.05 & 3.0 & 0.01 & 1.00 & 3.0 & 0.67 & 1.00 &  &  &  \\
Top-$|\widehat G|$ by energy & 0.05 & 0.05 & 3.0 & 0.01 & 1.00 & 3.0 & 0.67 & 1.00 &  &  &  \\
Both fits above the floor, no comparison & 1.00 & 1.00 & 4.3 & 0.27 & 1.00 & 4.1 & 0.76 & 1.00 &  &  &  \\
$\hat f_D$ only: $a_u$ above the floor & 1.00 & 1.00 & 5.0 & 0.37 & 1.00 & 5.1 & 0.80 & 1.00 &  &  &  \\
\midrule
\multicolumn{12}{l}{\emph{$p=30$}} \\
LENS (default) & 0.05 & 0.05 & 3.0 & 0.01 & 1.00 & 3.0 & 0.67 & 1.00 & 5.9 & 0.00 & 0.62 \\
$\hat\mu_u$ above the boundary, no test & 0.05 & 0.05 & 3.0 & 0.01 & 1.00 & 3.0 & 0.67 & 1.00 & 5.9 & 0.00 & 0.62 \\
Top-$|\widehat G|$ by energy & 0.05 & 0.05 & 3.0 & 0.01 & 1.00 & 3.0 & 0.67 & 1.00 & 5.9 & 0.00 & 0.62 \\
Both fits above the floor, no comparison & 1.00 & 1.00 & 5.8 & 0.48 & 1.00 & 4.0 & 0.75 & 1.00 & 6.3 & 0.04 & 0.64 \\
$\hat f_D$ only: $a_u$ above the floor & 1.00 & 1.00 & 6.0 & 0.50 & 1.00 & 4.5 & 0.77 & 1.00 & 6.5 & 0.06 & 0.64 \\
\midrule
\multicolumn{12}{l}{\emph{$p=50$}} \\
LENS (default) & 0.10 & 0.05 & 3.0 & 0.01 & 1.00 & 3.0 & 0.67 & 1.00 &  &  &  \\
$\hat\mu_u$ above the boundary, no test & 0.10 & 0.05 & 3.1 & 0.03 & 1.00 & 3.1 & 0.67 & 1.00 &  &  &  \\
Top-$|\widehat G|$ by energy & 0.10 & 0.05 & 3.0 & 0.01 & 1.00 & 3.0 & 0.67 & 1.00 &  &  &  \\
Both fits above the floor, no comparison & 1.00 & 1.00 & 5.9 & 0.49 & 1.00 & 4.0 & 0.75 & 1.00 &  &  &  \\
$\hat f_D$ only: $a_u$ above the floor & 1.00 & 1.00 & 6.1 & 0.51 & 1.00 & 4.2 & 0.76 & 1.00 &  &  &  \\
\bottomrule\end{tabular}\end{table}

\begin{table}[H]\centering
\caption{Selection rules on real data: mean list size and test loss over five splits (Poisson deviance for freMTPL2 at three training sizes per model; log loss for credit). Each list is refitted on top of the same additive model $f_0$.}
\label{tab:app-real}\footnotesize\setlength{\tabcolsep}{3.5pt}
\begin{tabular}{lrcrcrcrcrc}
\toprule
 & \multicolumn{6}{c}{freMTPL2 (rows per model)} & \multicolumn{4}{c}{Credit} \\
\cmidrule(lr){2-7}\cmidrule(lr){8-11}
 & \multicolumn{2}{c}{20,000} & \multicolumn{2}{c}{50,000} & \multicolumn{2}{c}{184,420} & \multicolumn{2}{c}{HELOC} & \multicolumn{2}{c}{UCI} \\
\cmidrule(lr){2-3}\cmidrule(lr){4-5}\cmidrule(lr){6-7}\cmidrule(lr){8-9}\cmidrule(lr){10-11}
Rule & size & loss & size & loss & size & loss & size & loss & size & loss \\
\midrule
Additive $f_0$ & -- & 0.31587 & -- & 0.31447 & -- & 0.31285 & -- & 0.55084 & -- & 0.43118 \\
Black box $\hat f_D$ & -- & 0.31747 & -- & 0.31002 & -- & 0.30536 & -- & 0.56082 & -- & 0.43097 \\
\midrule
LENS (default) & 6.6 & 0.31318 & 9.0 & 0.31168 & 12.0 & 0.30950 & 2.0 & 0.55076 & 5.2 & 0.43028 \\
LENS, $\kappa=0$ & 6.6 & 0.31318 & 9.0 & 0.31168 & 13.0 & 0.30945 & 23.2 & 0.55408 & 35.2 & 0.43177 \\
LENS, $b_u=\eta$ & 11.4 & 0.31340 & 14.8 & 0.31160 & 17.2 & 0.30938 & 55.4 & 0.55590 & 67.0 & 0.43266 \\
$\hat\mu_u$ above the boundary, no test & 6.6 & 0.31318 & 9.0 & 0.31168 & 12.2 & 0.30950 & 2.6 & 0.55038 & 5.4 & 0.43010 \\
Top-$|\widehat G|$ by energy & 6.6 & 0.31414 & 9.0 & 0.31251 & 12.0 & 0.31019 & 2.0 & 0.55066 & 5.2 & 0.43024 \\
$\hat f_D$ only: $a_u$ above the floor & 25.8 & 0.31382 & 26.4 & 0.31166 & 23.4 & 0.30921 & 30.2 & 0.55507 & 41.0 & 0.43173 \\
Both fits above the floor, no comparison & 22.2 & 0.31368 & 22.6 & 0.31164 & 21.6 & 0.30922 & 16.8 & 0.55345 & 15.4 & 0.43098 \\
All pairs & 36.0 & 0.31432 & 36.0 & 0.31189 & 36.0 & 0.30912 & 253.0 & 0.59148 & 253.0 & 0.44213 \\
\bottomrule\end{tabular}\end{table}

\textbf{The comparison removes what one fit learned.} Both rules without the
comparison return a pair in every run of the two additive designs, where every
pair is false; LENS does so in at most one run in ten
(Table~\ref{tab:app-ablation}). On Sparse their lists grow from 3.0 to 4.3--6.1
pairs with an FDP of 0.27--0.51, against 0.01 for LENS, and on F1--F10 their FDP
is 0.04--0.06, against 0, for a power gain of 0.02. Using the second model only
as a second threshold barely helps: \emph{both fits above the floor} is almost as
long as \emph{$\hat f_D$ only}, because a model fitted to noise leaves energy in many
pairs; the two fits agree on how strong a pair is, not on its pattern. The
same holds on real data (Table~\ref{tab:app-real}). On freMTPL2 the two rules
keep 22--26 of the 36 pairs, two to four times the LENS list. With 20,000
policies per model their loss is higher than that of LENS; with more data it is
lower by at most $3\times10^{-4}$, the gain of adding nearly every pair. On the credit data they keep 15--41
pairs, against 2.0 and 5.2, and their loss is higher than that of LENS on both
data sets; on HELOC it is higher than that of the additive model.

\textbf{The ranking alone cannot set the list.} Cut at the length of the LENS
list, the energy ranking matches LENS in size, FDP, and power on every synthetic
design: there the true pairs are also the strongest. The length, however, comes from
LENS, and on real data the strongest pairs are not always the reproduced ones. On
freMTPL2 the top-$|\widehat G|$ list has a higher test deviance than the LENS
list at every training size (0.31414 against 0.31318, 0.31251 against 0.31168,
and 0.31019 against 0.30950) and in each of the 15 splits. On the credit data
the two lists have the same loss to within $10^{-4}$.

\textbf{The test makes the list error-controlled at no cost in accuracy.}
Keeping every pair whose point estimate $\hat\mu_u$ clears the boundary gives
the LENS list in all but a few runs: it differs only at $p=50$ on Sparse and
Pair-corr (3.1 pairs each; FDP 0.03 against 0.01 on Sparse), on full freMTPL2 (12.2 pairs against 12.0), and on the credit
data (2.6 against 2.0 and 5.4 against 5.2), with losses within $4\times10^{-4}$ of
those of LENS. With the evaluation sets used here the standard errors are small, so the list is set by the boundary, as the insensitivity to $\alpha$
in Section~\ref{sec:exp-sensitivity} also shows. The pairs the point rule adds
are those within sampling error of the boundary, including the false pairs at
$p=50$ on Sparse. The test is what keeps them out and what gives the list the
error control of Theorem~\ref{thm:fdr}; the point rule has no such guarantee,
and nothing bounds its errors when the evaluation set is small.

\textbf{Summary.} The second model and the comparison remove the patterns that
only one fit learned, the boundary sets how long the list is, and the test
makes the list error-controlled. Removing the comparison brings back false pairs
on every design and longer, worse lists on real data; removing the test changes
almost nothing on these sample sizes but forfeits the guarantee.

\end{document}